\pdfoutput=1
\documentclass[11pt]{article}
\usepackage[final]{acl}

\usepackage{times}
\usepackage{latexsym}
\usepackage[T1]{fontenc}
\usepackage[utf8]{inputenc}
\usepackage{microtype}
\usepackage{tabularx}
\usepackage{inconsolata}
\usepackage{booktabs}
\usepackage{xspace}
\newcommand{\corpusname}{\textsc{Coping}\xspace}
\newcommand{\F}{F$_{1}$\xspace}
\usepackage{pifont}
\usepackage{graphicx}
\usepackage{multirow}
\usepackage{longtable}
\usepackage{markdown}
\usepackage{bm}
\usepackage{makecell}
\usepackage{colortbl}
\definecolor{lightgreen}{rgb}{0.76, 0.92, 0.73}
\colorlet{customgray}{lightgray!30}
\definecolor{pinkish}{HTML}{ED220D}
\colorlet{custompurple}{pinkish!30}

\title{Dealing with Controversy:\\ An Emotion and Coping Strategy Corpus Based on Role Playing}

\author{Enrica Troiano$^{1,2}$*, Sofie Labat$^3$*, Marco Antonio Stranisci$^{4,5}$*,\\\bfseries Viviana Patti$^4$, Rossana Damiano$^4$ \and Roman Klinger$^6$\\
$^1$Computational Linguistics and Text Mining Lab, Vrije Universiteit Amsterdam, Netherlands\\
$^2$HK3Lab, Rovereto, Italy\\
$^3$LT3, Language and Translation Technology Team, Ghent University, Belgium\\
$^4$Dipartimento di Informatica, Università degli Studi di Torino, Italy\\
$^5$aequa-tech, Turin, Italy\\
$^6$Fundamentals of Natural Language Processing, University of Bamberg, Germany\\
\texttt{\{firstname.lastname\}@\{hk3lab.ai,ugent.be,unito.it,uni-bamberg.de\}}
}

\begin{document}
\maketitle
\def\thefootnote{*}\footnotetext{The first three authors contributed equally.}
\renewcommand{\thefootnote}{\arabic{footnote}}
\begin{abstract}
  There is a mismatch between psychological and computational studies
  on emotions. Psychological research aims at explaining and
  documenting internal mechanisms of these phenomena, while
  computational work often simplifies them into labels. Many emotion
  fundamentals remain under-explored in natural language processing,
  particularly how emotions develop and how people cope with them. To
  help reduce this gap, we follow theories on coping, and treat
  emotions as strategies to cope with salient situations (i.e., how
  people deal with emotion-eliciting events). This approach allows us
  to investigate the link between emotions and behavior, which also
  emerges in language.  We introduce the task of coping
  identification, together with a corpus to do so, constructed via
  role-playing. We find that coping strategies realize in text even
  though they are challenging to recognize, both for humans and
  automatic systems trained and prompted on the same task. We thus
  open up a promising research direction to enhance the capability of
  models to better capture emotion mechanisms from text.
\end{abstract}

\section{Introduction}
\label{sec:introduction}
In the last decades, natural language processing (NLP) has examined
the interaction between language and emotions from different angles,
including text generation and
understanding~\cite{alhuzali-ananiadou-2021-spanemo,schmidt-etal-2021-emotion,gao-etal-2021-improving-empathetic,wang-etal-2022-empathetic}. This
work has resulted in the creation of many resources for processing
affect in text~\cite{bostan-klinger-2018-analysis}, largely based on
psychological theories that treat emotions as either discrete labels
or vectors in a space defined by the dimensions of valence, arousal
and dominance~\cite{Ekman1992,Plutchik2001,russell2012psychological},
enabling machine learning approaches to model emotions in the same
manner~\citep{buechel-hahn-2018-word,Demszky2020,hipson2021emotion}. To
date, however, much research disregards the complex mechanisms
governing the emergence and subsequent effect of emotions on behavior,
which are often conceptualized in psychology as coping
processes~\cite{lazarus1991emotion,Scherer1982}.  Essentially,
emotions can be considered episodes where people \textit{cope} with an
event, and that influence their behavior through action
tendencies~\cite{Lazarus1991}. According to \citet{Roseman2013}, this
influence occurs in four generic ways: they correspond to the four
coping strategies that people use to \textit{attack} an emotion
stimulus, \textit{contact} it, \textit{distance} themselves from it,
or \textit{reject} it, and each of them is contextual to feeling
specific emotions -- e.g., being disgusted by something aligns with
the tendency to reject it, feeling love prompts to establish contact
with the loved object.

\begin{table}[t]
\centering\small
\begin{tabularx}{\linewidth}{p{1cm}p{6cm}}
\toprule
Strategy & Example\\
\cmidrule(r){1-1}\cmidrule(r){2-2}
Attack & Every word you said just proves how utterly clueless you really are. I won't tolerate your ideas anymore.\\
Contact & Allow me to disagree. I feel like your ideas might have fatal repercussions.\\
\bottomrule
\end{tabularx}

\caption{Texts expressing different coping strategies.}
\label{Text-with-Coping-Example}
\end{table}

Theories of coping have great visibility in research on virtual agents
and robotics due to their ability to model emotions at the behavioral
level \cite[i.a.]{spekman2018perceptions,marsella2009ema}, but their
potential for studying emotions in language (as a specific type of
behavior) has not been fully explored so far. In this paper, we
propose to study textual emotions from the perspective of coping. As
NLP moves towards investigating emotions as phenomena fundamentally
centered on events \cite{Klinger2023a}, automatic systems are required
to understand how individuals cope with those events when feeling and
verbally expressing an emotion. They thus urge us to question the way
that coping strategies realize in language, and if they can be
captured by computational models.

We hypothesize that these strategies, as essential building blocks of
communication, emerge in text just like the emotions that they
accompany (anger, hope, etc.). Intuitively, a text expressing anger
should also convey the intention of attacking a stimulus, and one
associated to hope might verbalize a contacting-type of coping (see
Table~\ref{Text-with-Coping-Example} for some examples).  Our study
sets out to verify this assumption: we construct \corpusname (COPIng
emotioN strateGies), the first corpus focused on coping information,
and we investigate the feasibility of the task of coping strategy
detection.

We approach this task using dialogues that depict conflictual
situations about topics of actual relevance, in which one interlocutor
utters a controversial statement threatening one's identity, and the
other person fights against it while implementing a coping strategy.
In ecological settings (e.g., on online platforms), finding
interactions of this type annotated with coping information is hard.
Hence, we gather them via crowdsourcing, building \corpusname with an
experimental design for text generation and annotation based on
role-playing.  Role-play provides a methodologically founded way to
elicit behavioral and linguistic responses by asking individuals to
put themselves in the shoes of a fictional (yet credible) character,
with whom they do not necessarily share concerns and values. In our
case, people impersonate a character that portrays one of Roseman's
coping strategies, and produce text to disagree with an imaginary
interlocutor.

Using \corpusname, we answer the following research questions.  (RQ1)
Can coping strategies be elicited in language and subsequently
annotated?  With a quantitative and qualitative analysis of the
corpus, we show that this is the case -- with some caveats deriving
from our data collection procedure.  (RQ2) Can coping strategies be
identified automatically?  To answer this question, we fine-tune
models for the classification of coping strategies on \corpusname, and
compare them to humans and four prompt-based Large Language Models
(LLMs) performing the same task in a zero-shot setup.

In sum, our contribution is threefold. (1) We take an initial step to
study the four coping categories that underlie the labels typically
used in computational emotion analysis, driven by the taxonomy of
attack, contact, distance, and reject that Roseman links to $16$
emotions (see Table~\ref{tab:coping-definitions}). (2) We release a
highly-structured corpus, built with a novel strategy for the
collection of emotion-centered information. (3) We introduce and
benchmark the task of coping strategy identification. Experimental
results support its viability.

Our corpus and code are available at
\url{https://www.uni-bamberg.de/nlproc/ressourcen/emotioncoping}.

\section{Related Work}

\begin{table}[t]
  \centering\small
\begin{tabular}{p{1.4cm}p{2.5cm}p{2.5cm}}
\toprule
{\bf Coping Strategy} & \mbox{}\par{\bf Emotions} & {\bf Behavioral\par Function}\\
\midrule
Attack & Anger, frustra-\newline tion, guilt & Move against \newline stimuli\\
\cmidrule(r){1-1}\cmidrule(r){2-2}\cmidrule(r){3-3}
Contact & Joy, hope, love, pride, relief & Increase contact and interaction\newline with stimuli\\
\cmidrule(r){1-1}\cmidrule(r){2-2}\cmidrule(r){3-3}
Distance & Dislike, distress, fear, regret, sadness & Decrease \newline contact and \newline interaction with stimuli\\
\cmidrule(r){1-1}\cmidrule(r){2-2}\cmidrule(r){3-3}
Reject & Contempt, dis-\newline gust, shame & Move stimuli\newline away\\
\bottomrule
\end{tabular}
\caption{The four coping strategies with the corresponding emotion groups and behavioral functions, as in~\citet{Roseman2013}.}
\label{tab:coping-definitions}
\end{table}

We now summarize the concepts of coping and emotions, and related NLP
research.

\subsection{Coping Strategies and Emotions}\label{sec:psych_theory}
According to well-established psychological theories, emotions emerge
as a person evaluates, or appraises, salient
situations~\cite{Scherer2001}. For example, if someone receives a
critical review, they might appraise it as a sign of failure and feel
anxious, or appraise it as constructive and feel determined to
improve. Within this theoretical framework,~\citet{Roseman2013} links
emotions to coping, a process useful for ``responding to situations of
crisis and opportunity'' (p.143).

More precisely, emotions are instances of the general strategies of
\textit{attack}, \textit{contact}, \textit{distance}, and
\textit{reject} (as in Table~\ref{tab:coping-definitions}) that serve
to seek more or less of a stimulus. For example, emotions like anger
and frustration correspond to the \textit{attack} strategy by which
people engage with a source of distress; hope and joy pertain to the
\textit{contact} strategy, which increases one's interaction with the
stimulus; fear and sadness belong to the \textit{distance} strategy,
that minimizes the source of distress; disgust and shame share a
\textit{reject} strategy to eliminate the stimulus.

Roseman's theory is accompanied by others on the same matter, such as
that of \citet{lazarus1984stress}, which sees coping as two
simultaneous stages: (1) a person evaluates the motivational relevance
and congruence of a situation in respect to their own goals: ``How
relevant and (in)consistent is the situation with my own
needs/goals?''~\cite{Smith2009}; and (2) the situation is evaluated in
terms of coping potential: ``Which resources and options do I have for
coping?''~\cite{Smith2009}. This theory further distinguishes between
problem-focused and emotion-focused coping. Problem-focused strategies
are directed at the stimulus (e.g., decreasing environmental
pressures) or at the self \cite[e.g., altering level of
aspiration,][]{kahn1964}, while emotion-focused coping involves
cognitive processes that reduce emotional distress (e.g., avoidance or
acceptance) or increase it to mobilize for
action~\cite{lazarus1984stress}.  While the distinction between
problem- and emotion-focused coping was criticized in later research
for being non-exclusive and non-exhaustive~\cite{skinner2003}, the
adoption of a theory that relies on the interaction between coping
mechanisms and the corresponding emotions is promising to explore more
in-depth computational approaches to emotion detection in NLP.  We
therefore follow the theory of~\citet{Roseman2013}, which provides an
exhaustive mapping of this type.

\subsection{Psychological Theories and NLP}
\label{sec:nlp}
The majority of works in NLP have used discrete emotion categories to
extrapolate from text, borrowed from theories such as Ekman's
(\citeyear{Ekman1992}) and Plutchik's (\citeyear{Plutchik2001}). There
have also been successful efforts at identifying the underlying
components of emotions, in terms of valence, dominance, and
arousal~\cite{mohammad-2018-obtaining,Park2021}, following dimensional
models of affect~\cite[e.g.,][]{russell2012psychological}. Our work,
however, is closer to a more recent research direction, which
considers the interplay between emotions and their eliciting events to
carry out a discussion and the modeling of
appraisals~\cite{yeo-jaidka-2023-peace,zhan-etal-2023-evaluating}.

\begin{figure*}
\centering
\includegraphics[width=14cm,height=3.5cm]{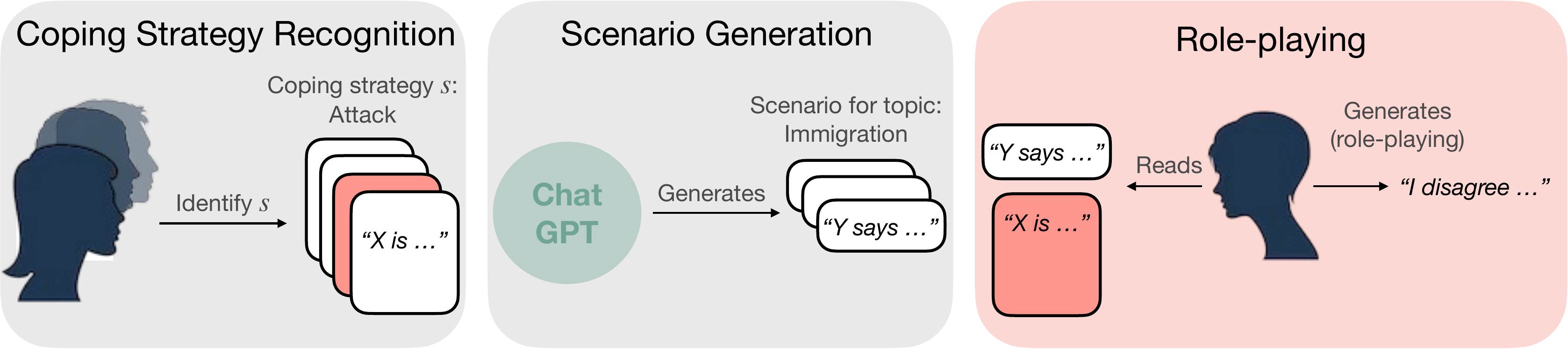}
\caption{Crowdsourcing methodology. Grey boxes: collecting the
  building blocks for role-playing. Annotators identify the coping
  strategy $s$ (in the figure, attack) in the description of character
  $X$; separately, ChatGPT-4 produces scenarios for different topics
  (in the figure, immigration).  Pink box: role-playing. A character
  description and a scenario on a given topic are presented to
  annotators as textual prompts to induce coping-related text
  generation.}
\label{fig:methods}
\end{figure*}

Related to that, some NLP works investigate
re-appraisal~\cite{Uusberg2019}, i.e., how people re-interpret a
situation to alter its emotional impact (e.g., ``When I realized I
could listen to my podcast, I stopped feeling frustrated about the
traffic jam.'').  For instance, based on re-appraisal,
\citet{li_herderich_goldenberg_2024} and~\citet{ziems2022} reframe
texts that depict negative situations in a more positive manner.
Similarly, we focus on emotion responses to data evoking adverse
circumstances, but we use verbal stimuli that can distress the
annotators (by describing scenarios with contentious interactions) in
order to activate coping (similar to~\citet{Saha2019}
and~\citet{Choi2020}).  In doing so, we address the interplay between
emotions and events via coping, providing the first corpus annotated
with this phenomenon and exploring the effective emergence of
Roseman's strategies in text.

\section{Data Creation Methodology}
Textual data associated with coping strategies is hard to
obtain. Ideally, since coping is a response to an appraised situation,
both the situation itself and the textually expressed coping strategy
should be known -- but the former may be external to the text and
impossible to track.  Moreover, coping strategies rarely appear as
naturally-occurring labels of texts (e.g., as hashtags assigned by
writers), and they involve complex concepts which might be unfamiliar
to laypeople (i.e., the psychological sense of {\it attack, contact,
  distance} and {\it reject}).

We solve these issues by resorting to a role-playing approach inspired
by the paradigm of improvisational theater~\cite{baumer2009narrative}
and supported by research on character
identification~\cite{giovannelli2009sympathy,battaglino2014moral}:
instead of annotating existing data, participants engage in a
make--pretend activity in which they write new texts while
(purposefully) instantiating a given coping strategy.\footnote{Workers
  are recruited on Prolific, \url{https://www.prolific.com}.} We
approximate the collection of data from first-hand experiencers of
coping.

We apply this approach as in Figure~\ref{fig:methods}.  We start by
gathering the building blocks of the role-playing study
(Section~\ref{ssec:describing-coping}), namely, definitions for the
four strategies and texts depicting scenarios in which these might be
triggered.  Based on such definitions and scenarios, we stimulate the
production of coping-expressing utterances
(Section~\ref{ssec:role-play-annotation}).

\subsection{Building Blocks of Role-playing}
\label{ssec:describing-coping}
\begin{table*}
\centering
{\small
\begin{tabular}{p{1.8cm}p{12.8cm}}
\toprule
\rowcolor{customgray} Definition & This character comes across as a calm, understanding, and very approachable person. For X, communication serves to unite people. It is an opportunity to exchange opinions, acknowledging the diversity of perspectives among individuals. When problems or unpleasant situations arise, this character responds with a constructive attitude. X expresses ideas with confidence, trying to solve problems in a respectful manner. X can effectively engage in discussions also with people having contrasting opinions.\\
\cmidrule(r){1-1}\cmidrule(r){2-2}
\rowcolor{customgray}Scenario \newline (topic: racism) &During a university class discussion on historical racial events, Y confidently states, ``People keep talking about systemic racism, but I believe that's just an excuse for those who don't want to work hard. If you look around, everyone has the same opportunities today.''\\
\midrule
\rowcolor{custompurple}Generated \newline reply & I understand your point Y, but it is not the case for everyone. Our group is a select handful of people who have been brought up this way.\\
\cmidrule(r){1-1}\cmidrule(r){2-2}
\rowcolor{custompurple}Additional \newline annotations & Description of X's non-verbal behavior; rating of X's emotional responses; comparison with own reaction.\\
\bottomrule
\end{tabular}
}
\caption{Example description of $X$ (i.e., definition for the coping strategy contact), with a scenario for the topic racism, and a text generated by a participant presented with the former, during the role-playing phase. The text is complemented with information on $X$'s behaviors and emotions, and the participants self-reflection on their own.}
\label{tab:exp-build-blocks}
\end{table*}

Good definitions of the strategies under consideration retain
Roseman's insights and minimize his technical jargon.  We thus
formulate each strategy $s$ as the profile of a fictional character
called $X$, who has all idiosyncrasies of that $s$.  We use these
descriptions\footnote{Descriptions are in
  Appendix~\ref{sec:data-creation-descriptions}, with guidelines
  (Table~\ref{tab:pre-study-descriptions-4}) and results
  (Figures~\ref{fig:confusion-coping-pretest} and
  ~\ref{fig:spiders-emotions-pretest-round3},
  Tables~\ref{attack-contact-emo-pretest-round4} and
  ~\ref{distance-reject-emo-pretest-round4}).}  to crowdsource the
task of coping strategy recognition, where people infer a coping
strategy from a given description of $s$. This labeling task $t$ can
be conceptualized as estimating p($s\mid\text{description}(s)$), and
it serves as a pretest for the subsequent crowdsourcing phase of text
generation and annotation
(Section~\ref{ssec:role-play-annotation}). The rationale is that the
annotators' performance will signal the quality of the description:
the more a description is mapped to the correct $s$, the clearer it
is.

As for the scenarios, we start from the idea that coping occurs in
moments of crisis and opportunity. Hence, we identify five topics of
current social relevance which might spur emotions and personal
stances, i.e., abortion, immigration, racism, LGBTQ+ rights, and
drugs. We generate three controversial scenarios for each of them, in
the form of fictional dialogical settings where an interlocutor puts
forward a contentious argument.
An example of a coping definition and scenario is in Table~\ref{tab:exp-build-blocks}.

\paragraph{Coping Strategy Recognition: Pretesting $\bm{X}$'s Descriptions.} 
Our goal is to observe if crowdworkers associate $X$ with the correct
coping strategy: presented with a description of $X$, the annotators
pick one of four possible strategies. We conduct this task in four
rounds, each time modifying both the descriptions of $X$, and the set
of possible answers presented to the annotators.\footnote{ Initially,
  the four answer options comprised the coping strategy labels (e.g.,
  {\it attack}); later, we opted for one-line summaries of $X$'s
  profile -- e.g., {\it X directly approaches problems with a
    potentially conflicting attitude (i.e., $X$ is an attacking-kind
    of person)}.}  From the candidate descriptions collected in all
rounds, we select the four that elicit the highest annotation accuracy
(attack and contact: 100\% participants recognized the correct
strategies, distance: 92\%, reject: 52\%). We use those to write the
guidelines for the role-playing task of
Section~\ref{ssec:role-play-annotation}.

\paragraph{Scenario Generation.} The scenarios are gathered via
ChatGPT-4 to mitigate our personal and cultural stances.\footnote{All
  prompts for ChatGPT-4 are in
  Appendix~\ref{sec:appendix-coping-scenario},
  Table~\ref{tab:scenarios-chat}; the generated scenarios for each
  topic in Table~\ref{tab:scenarios-per-topic}.} We ask the LLM to
produce three descriptions of every-day situations (e.g., meet-ups in
a bar), in which participants $X$ and $Y$ interact, and $Y$ utters a
strong opinion on a sensitive topic. ChatGPT-4 is required to specify
the power status between $X$ and $Y$ (i.e., they are either two peers,
or have a power imbalance), as we assumed that the way people respond
to $Y$ could depend on what relationship they hold with such
interlocutor.

\subsection{Role-Playing}
\label{ssec:role-play-annotation}

We obtain texts associated with coping strategies by crowdsourcing
task $t'$ = p(generation $\mid t$): participants\footnote{These
  workers do not overlap with those in
  Section~\ref{ssec:describing-coping}.} generate text based on their
understanding of a description of $s$. Their assignment is not only to
form a mental representation of $X$ and write a text: they must
impersonate $X$ and write as $X$ would, in such a way that the
resulting text discloses $X$'s personal features.  Since $X$
represents a coping strategy, we assign the texts so collected to the
corresponding $s$.

In practice, we collect the texts with $60$ questionnaires, each
mapping to a combination of a description of $X$ and a scenario on a
given topic. All questionnaires ask participants to map the
description of $X$ to a short summary of $X$'s profile (as in the
previous task $t$). After the crowdworkers familiarize themselves with
$X$, they read the scenario that includes the hostile interlocutor
$Y$. Next, as sketched in Table~\ref{tab:exp-build-blocks}, they: (i)
generate a reply that opposes $Y$'s statement by pretending they were
$X$; (ii) describe $X$'s non-verbal reaction; (iii) rate the emotional
response of $X$ on a scale from $1$ to $5$ for each emotion in
Roseman's taxonomy. (iv) Lastly, participants are asked if they
themselves would react verbally and non-verbally like $X$. If their
answer is negative, they repeat the first three steps from their own
perspective instead of $X$'s.\footnote{Details on questionnaires and
  participants in Appendix~\ref{sec:appendix-role-play}.} The 60
questionnaires are administered consecutively, with later versions
featuring slight adjustments to the instructions for clarity, while
maintaining consistency in both the tasks and the flow of questions.

We anticipated that responses to tasks (i) to (iv) might be influenced
by both the participants' role-playing abilities and their own
involvement in the topic under consideration. Hence, to evoke bolder
opinions, potentially resulting in clearer coping strategy
expressions, we select participants with specific demographic features
and assign them to a topic accordingly (e.g., LGBTQ+ scenarios are
given to crowdworkers that belong to this community, and racism-based
scenarios are presented to people self-assigning to an ethnic
minority).

\section{The \corpusname Corpus}
\label{sec:data-statistics}

\begin{figure*}[t]
\centering
\begin{tabular}{cccc}
Attack & Contact & Distance & Reject \\
\includegraphics[width=.22\textwidth]{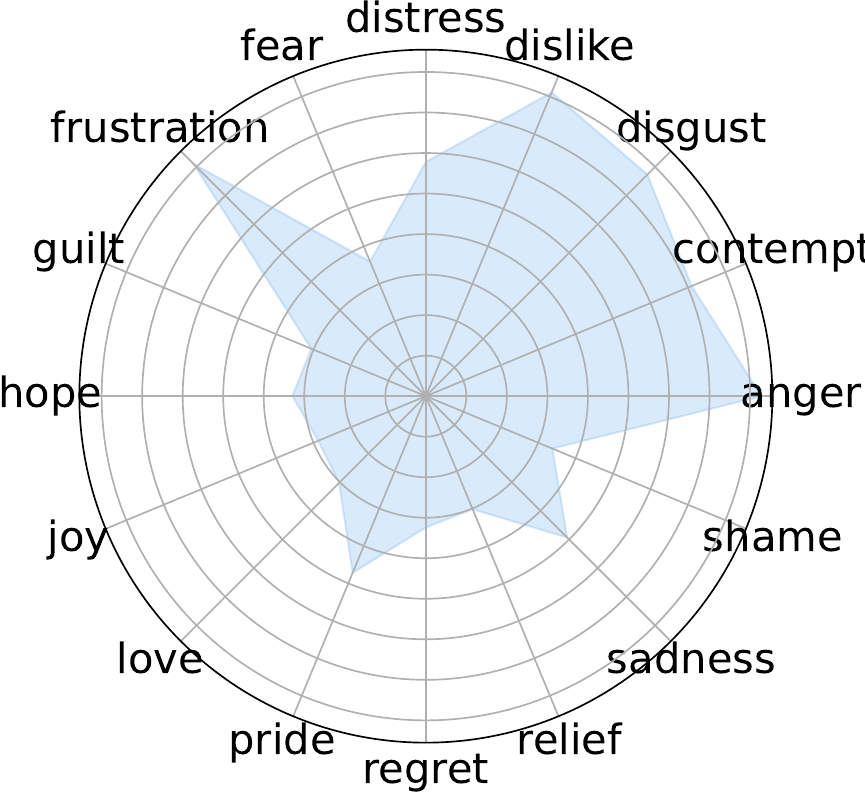}
& 
\includegraphics[width=.22\textwidth]{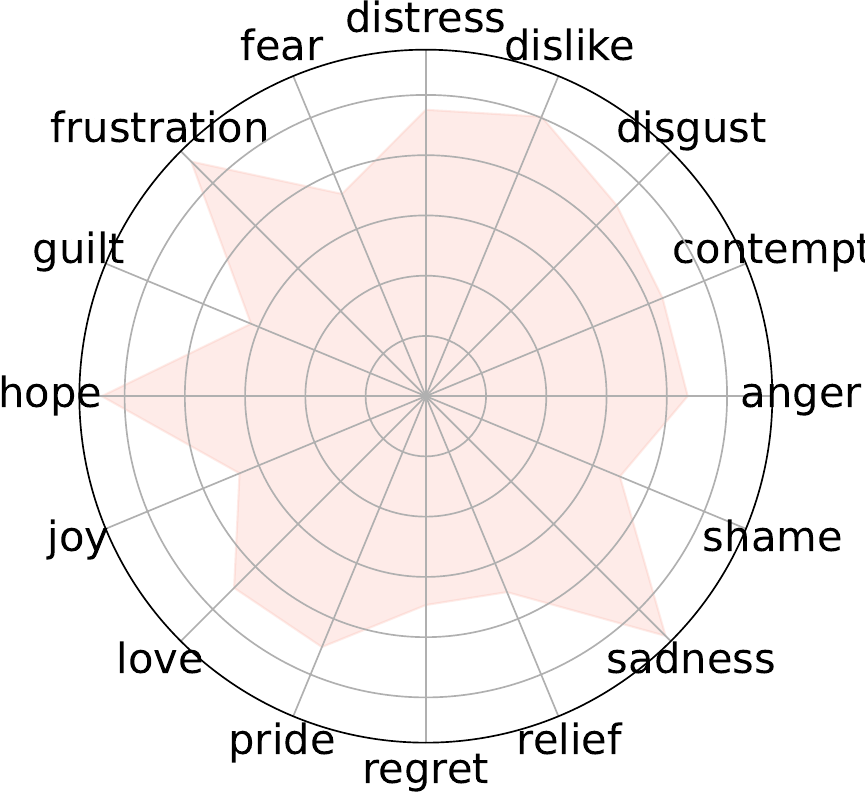}
&
\includegraphics[width=.22\textwidth]{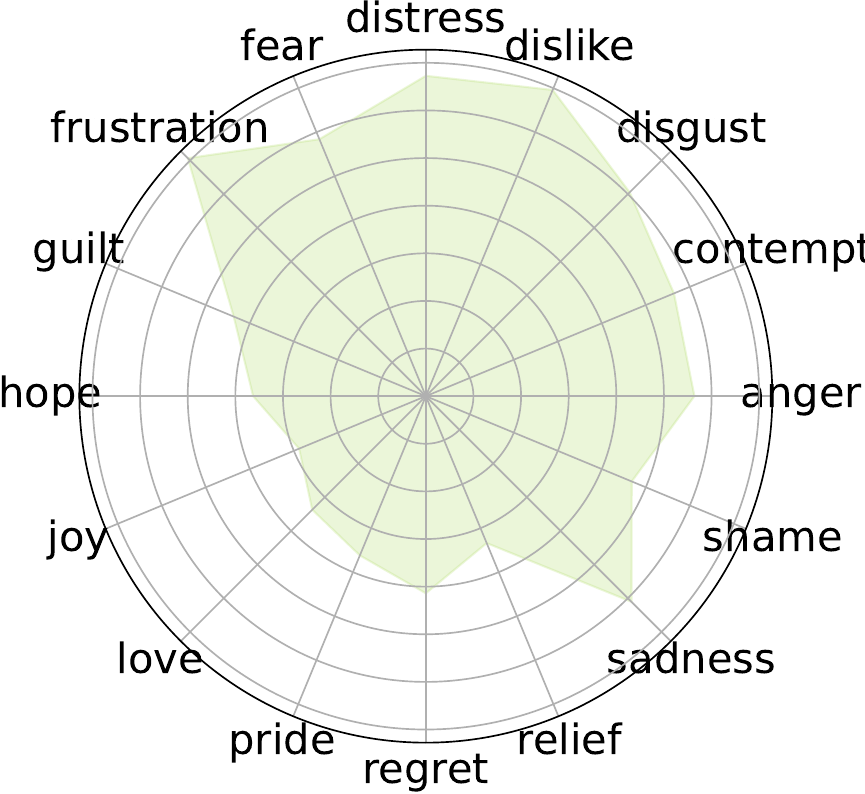}
&
\includegraphics[width=.22\textwidth]{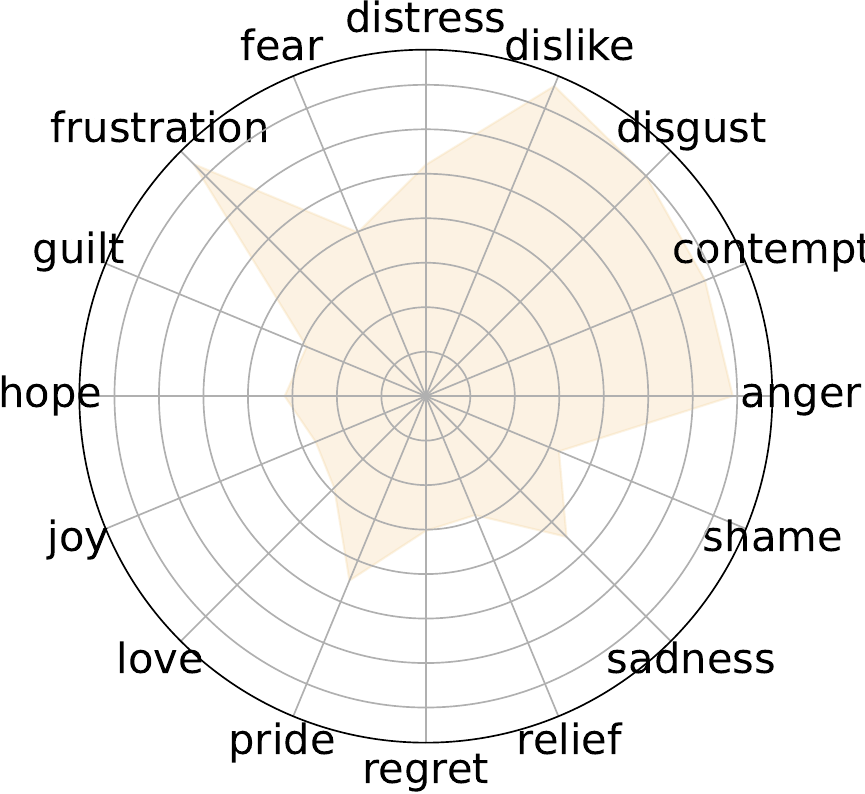}
\end{tabular}
\caption{Emotion averages of character $X$ for each coping strategy.}
\label{fig:emo-av}
\end{figure*}
\corpusname contains $1,200$ data points, balanced among coping
strategies, topics, and scenarios. On average, the length of $X$'s
answer to $Y$ is $34.8$ tokens\footnote{Tokenization performed with
  spacy \texttt{en\_core\_web\_sm}.}, that of texts about $X$'s
behaviors is $23.6$. We now analyze the corpus to investigate if the
four descriptions of $X$ (such as that in
Table~\ref{tab:exp-build-blocks}) allow for a correct understanding of
coping strategies. In doing so, this section answers \textbf{(RQ1) Can
  coping strategies be elicited in language and subsequently
  annotated?} We investigate coping understanding from different
angles. First (Section~\ref{ssec:writers}), we do this irrespective of
how the understanding reflects in text -- to separate such an
understanding from people's ability to write utterances while
role-playing.  Then (Section~\ref{ssec:text-analysis}), we question
how well texts signal the presence of specific coping strategies.

\subsection{Coping from the Perspective of Writers}
\label{ssec:writers}
Our work adheres to Roseman's framework, but does so in an
``artificial'' experimental setup. We therefore test if the emotion
judgments provided by participants are theoretically expected.  If
participants correctly impersonated $X$ (for a given $s$), that has to
reflect in a distribution of emotions similar to Roseman's (cf.\
Table~\ref{tab:coping-definitions}) -- attack, e.g., should be
associated to a high level of anger, frustration, and guilt.  Further,
we observe how such emotions are different from the annotators' own
emotions.

\paragraph{Emotions of Character $\bm{X}$.} 
Figure~\ref{fig:emo-av} illustrates the distribution of emotions per
coping strategy.  Different coping strategies are characterized by
different emotion configurations: the mapping between Roseman taxonomy
and the judgment of the crowdworkers is correct for most emotions
(e.g., frustration, hope, dislike, contempt for attack, contact,
distance, and reject respectively). There are exceptions, however.
For example, crowdworkers failed to associate guilt to attack, shame
to reject, and love and relief to contact.  Further, they assigned
high values to, e.g., frustration in contact and anger in distance:
while theoretically imprecise, such scores can be an effect of the
scenarios presented to the crowdworkers, based on a hostile
interlocutor and a communication setting that presumably induce
negative emotions in $X$.

We corroborate these insights through correlations between emotions
and coping strategies.  We compute the Pairwise Cram{é}r's V
association~\cite{cramer1999mathematical} between the two variables,
and deem them correlated only when $V > .3$ (i.e., moderate to strong
correlations).  To decide which coping strategy an emotion is most
strongly associated with, we consider its average score, and assign it
to the coping strategy where that average is the highest -- e.g.,
anger correlates with attack because $V > .3$ and the average
respondents' rating is highest on that strategy ($4.14$) than any
other.  Results are in Table~\ref{tab:emotions-correlation}, where the
emotion patterns underlying each strategy suggest that the
crowdworkers achieved a sufficient level of understanding of character
$X$.  Six emotions (marked in bold) prove coherent with Roseman's
framework. For instance, frustration correlates with attack, hope with
contact, fear with distance.

Vice versa, lacks of correlations for joy, pride, regret, relief, and
sadness indicate that these emotions were not well identified by the
participants. Rather than a negative outcome, we take this observation
as an indicator of the success of role-playing, which made positive
emotions such as joy and pride unlikely for $X$.  Unsurprisingly,
reject turns out uncorrelated to any emotion, since those documented
by Roseman (i.e., contempt, disgust, and shame) are (incorrectly) more
linked to other coping strategies. This outcome points not only at the
difficulty for respondents to identify reject
(cf.\ Section~\ref{ssec:describing-coping}), but also at the fact that
such emotions are interwoven with others. For instance, in
Table~\ref{tab:emotions-correlation}, shame falls under the same
coping strategy as guilt -- both are indeed related to
self-forgiveness~\cite{leach2017understanding}; disgust, like anger,
is a response to a moral violation~\cite{molho2017disgust} and part of
the so called CAD triad (Contempt, Anger and Disgust) of moral-based
emotions according to~\citet{rozin1999cad} -- the three are often
elicited in combination in hostile
situations~\cite{matsumoto2016effects}, which might explain why they
are all correlated with attack.

Therefore, even if not perfectly aligned with Roseman's taxonomy, the
workers' understanding of coping appears meaningful, based on both
theoretically-established links between emotions and our experimental
setup for data collection.

\begin{table}
    \centering\small
    \begin{tabular}{lp{5cm}}
    \toprule
    Strategy & Emotion \\
    \cmidrule(r){1-1}\cmidrule(r){2-2}
      attack & \textbf{anger frustration} contempt\par disgust dislike    \\
      contact & \textbf{hope love}   \\
      distance   & \textbf{distress fear} shame guilt\\
      reject   & --\\
        \bottomrule
    \end{tabular}
    \caption{Coping strategies and correlated emotions (Cramer's $V>.3$). Emotions included in Roseman's framework for a given strategy are marked in bold.}
    \label{tab:emotions-correlation}
\end{table}

\paragraph{Effect of Role-Playing.} 
To further evaluate the participants' understanding of coping, we now
observe if they disentangled $X$'s emotions from their own. We focus
on the answers of the $571$ respondents ($47.5\%$ of the crowdworkers)
who claimed that they would not react as $X$, and we test this
self-assessment by performing a~\citet{Wilcoxon1945} signed-rank test
between the emotion scores provided while role-playing $X$ and the
scores provided from their personal perspective (e.g., between the
scores of $X$'s anger and the participant's). We find that claiming a
different personal reaction corresponds indeed to different emotion
reactions: except for pride (p-value: $.06$), all pairs of values for
the same emotion are significantly different.\footnote{A comparison
  between $X$'s emotions and the respondents' self-assigned emotions
  is in Appendix~\ref{sec:appendix-corpus-analysis}.} In sum, $X$’s
emotions and the annotators’ emotions can be clearly separated. This
corroborates that the participants successfully engaged in the
make-pretend activity, distinguishing their affective responses from
those imagined for the fictional character.

\subsection{Coping from the Perspective of Readers}
\label{ssec:text-analysis}
Focusing only on the textual responses gathered with role-playing, we
now reverse the direction of inquiry, and question if the texts
express the intended coping strategies according to readers.

\paragraph{Inter-Annotator Agreement.}
We conduct an in-house annotation study to see if said strategies are
successfully recognized by external people.  We randomly select $100$
texts and have $3$ annotators decide if and what coping strategy
emerges from each of them. Inter-annotator agreement computed
with~\citeauthor{fleiss1971measuring}' kappa
(\citeyear{fleiss1971measuring}) shows that the annotators' judgments
were not consistently similar ($\kappa=.34$). Annotators achieved the
highest agreement on the strategy of contact (.60) and the lowest on
reject (.04).  On $42$ texts, they provided unanimous
judgments. However, on only $28$ of these items were the correct
prompting labels recognized (mostly, attack and
contact). Surprisingly, a majority vote was reached on only a few more
texts ($45$); that vote matches the correct label for $18$ of
those. This finding hints at the fact that the intended coping
strategies are recognized when they are as clear as to put many people
in agreement.

Annotation differences were potentially influenced by the interaction
between coping strategies and topics, with certain topics making the
expression of the strategies harder for the crowdworkers, and
consequently limiting the $\kappa$ values among the readers. Overall,
like emotions, whose recognition can lead to low inter-annotator
agreement, the perception of coping in text proves extremely
subjective as well. Accordingly, the adoption of psychological
theories for the annotation of NLP resources proves a challenging task
deserving systematic analysis.

\begin{table}
    \centering\small
    \begin{tabular}{lp{5.8cm}}
    \toprule
    Strategy & Tokens\\
    \cmidrule(r){1-1}\cmidrule(r){2-2}
        attack & you, wrong, stupid, dare, racist\\
        contact & understand, help, feel, while, however\\
        distance & I, subject, sure, but, opinion\\
        reject & do, not, negative, want, attention\\
    \bottomrule
    \end{tabular}
    \caption{The five tokens most strongly associated with each strategy.}
    \label{tab:jensen-shannon}
  \end{table}
  
\paragraph{Differences in Linguistic Patterns.}
To better understand these annotation results, we look for significant
forms of variation between texts in \corpusname pertaining to
different coping strategies.  For each strategy, we split \corpusname
in two parts: the first with all the texts related to the given
strategy, and the second with all the remaining texts. We then compare
the distribution of tokens in the two sets, by computing the
Jensen--Shannon divergence~\cite{menendez1997jensen} for each token.

Table~\ref{tab:jensen-shannon} lists the five tokens most strongly
associated with each strategy. A set of tokens appears highly coherent
with attack (e.g., wrong, dare) and contact (e.g., understand, feel).
A manual qualitative analysis of the texts corroborates this finding.
Contact might be relatively easy to recognize thanks to politeness
markers (e.g., ``\textit{Allow me to disagree}''), while the detection
of attack corresponds to opposite linguistic strategies, such as
pressing questions (``\textit{How are they creating chaos by just
  existing?}''), direct confrontations (``\textit{Only a fool wouldn't
  not see the obvious}''), and expressions of how strongly $X$ feels
about their opinion or the interlocutor's (e.g., ``\textit{I could not
  disagree more}'', ``\textit{you are completely wrong}'').

Specific linguistic patterns are less evident for distance and reject,
associated with tokens that vaguely express them. The difference with
the other two coping strategies might be explained referring to
Roseman's theory. Attack and contact require interaction with the
emotion stimulus; distance and reject disengage from it. Role-playing
disengagement might be more challenging, because it requires
participants to react to a stimulus while detaching themselves from
it. Ultimately, these observations point out that some coping
strategies are easier to be imbued (and recognized) in text, at least
when they are purposefully produced.

\section{Coping Strategies Classification}
\textbf{(RQ2) Can coping strategies be identified automatically?} We
answer our second research question using fine-tuned and prompt-based
language models for coping strategy detection.

\begin{table}
    \centering\small
    \begin{tabular}{lrrr}
    \toprule
         & \textbf{Answer} & \textbf{Behavior} & \textbf{Answ.+Behav.} \\
         \cmidrule(r){2-2}\cmidrule(r){3-3}\cmidrule(r){4-4}
        BERT & $.401$ & $.395$ & $.457$\\
        DistilBERT & \textbf{.523} & $.443$ &  $.507$ \\
        RoBERTa & $.477$  & \textbf{.537} & \textbf{.512} \\
        \midrule
        phi\-2 &.356 & .391& .322\\
        mistral\-2& .392& .474&{\bf .538}\\
        dolphin\-2& {\bf .422}& .{\bf 550} &.520\\
        nous-hermes& .387&.526 &.515\\
        \bottomrule
    \end{tabular}
    \caption{Top: \F scores of the SLMs ran on $X$'s answers (Answer),
      $X$'s behavior (Behavior), and a concatenation of both
      (Answ.+Behav.). Bottom: \F scores obtained by feeding the four
      LLMs (rows) with a concatenation of instructions and input text
      (Answer, Behavior, both).}
    \label{tab:Bert-and-others}
\end{table}

\paragraph{Models.}
We fine-tune three Small Language Models (SLMs) on the \corpusname
texts: BERT~\cite{devlin-etal-2019-bert},
DistilBERT~\cite{sanh2019distilbert}, and
RoBERTa~\cite{liu2019roberta}. The fine-tuning is performed on three
separate sets of texts: $X$'s answers, $X$'s behaviors, and their
concatenation. After splitting the corpus in a train and a test set
(90\% and 10\% of the data), we fine-tune each model over three random
train/validation splits and train it for $10$ epochs with a batch size
of $10$.\footnote{Each training has been performed on a NVIDIA
  RTX-3060 Ti and lasted $25$ minutes on average.} As for the
prompt-based models, we implement zero-shot classifiers with four
open-source LLMs, i.e., phi-2, mistral-2, nous-hermes-2, dolphin-2,
reproducing the design used with SLMs (answer, behavior, and their
concatenation) on the same test
data.\footnote{Appendix~\ref{sec:appendix-modeling} details models,
  prompts, and full results.}

\paragraph{Results.}
The classification of coping strategies turns out challenging, both
for the SLMs and the prompt-based models. As shown in
Table~\ref{tab:Bert-and-others}, which reports the best \F score
obtained by the classifiers on the test set, all results are below
$.55$.

We observe differences for different types of input texts. DistilBERT
has the best classification performance when fed with answers
(\F=$.523$), while RoBERTa performs better in the classification of
$X$'s behaviors, with a score of $.537$. The combination of replies
and reactions does not lead to a better performance: the best model is
RoBERTa with an \F score of $.512$, which is lower than the ones
obtained for replies and reactions separately. The zero-shot
classifiers are on par with the others. Once more, the outputs based
on behavior alone surpass the two other setups. Here, the difference
between Answer and Behavior is much more noticeable (12 points in \F
for dolphin-2, $\approx$14 \F points for nous-hermes).

Zooming in the results by topic (Table \ref{tab:dis_cop}), we see that
both answers ($.623$ \F score) and behaviors ($.630$) in scenarios
related to abortion are better recognized by the fine-tuned DistilBERT
(answers) and RoBERTa (behaviors). The lowest \F score on $X$'s
answers is obtained with immigration-related scenarios ($.365$), while
the worst classification of $X$'s behaviors with racism-related
scenarios ($.457$).  We also notice that not all coping strategies are
equally difficult to detect (cf.\ Table~\ref{tab:dis_cop}). The
easiest is contact, classified with an \F score of $.647$ using $X$'s
answers and $.617$ using behaviors. Conversely, the classification of
reject results in the lowest \F scores: $.408$ for answers and $.428$
for behaviors.  This aligns with the human recognition of coping
strategies, with contact and reject defining the best and worst
recognized labels using character $X$'s answers (cf.\
Section~\ref{ssec:text-analysis}).

\begin{table}
    \centering\small
    \begin{tabular}{rlrr}
    \toprule
        && \textbf{Answer} & \textbf{Behavior}\\
        \cmidrule(r){3-3}\cmidrule(r){4-4}
        \multirow{5}{*}{\rotatebox{90}{Topics}} & Abortion & .623 & .630\\
        & Drugs & .513 & .488\\
        & Immigration & .365 & .483\\
        & LGBTQ+ & .508 & .619\\
        & Racism & .570 & .457\\
        \midrule
       \multirow{4}{*}{\rotatebox{90}{Labels}} &Attack & .500 & .545\\
        &Contact & .647 & .617\\
        &Distance & .539 & .560\\
        &Reject & .408 & .428\\
        \bottomrule
    \end{tabular}
    \caption{Classification results (\F) broken down by topic and coping
      strategy. Answers have been classified with DistilBERT,
      behaviors with RoBERTa.}
    \label{tab:dis_cop}
\end{table}

Put together, these numbers suggest that $X$'s answers provide some
coping signal, but the texts about $X$'s behavior are key to capture
the ``seeking more of an event'' or ``seeking less of it'', central to
the emergence of coping and its link with behavior.

\section{Discussion and Conclusion}

This paper follows a recent wave of
studies~\cite[i.a.]{cortal-etal-2023-emotion,liu-jaidka-2023-psyam}
that initiated a dialogue with other disciplines (like psychology), to
broaden the tools of computational emotion analysis and push forward
its potential to shed light on how emotions work in language. We
focused on the entanglement between emotions and coping strategies,
never-before explored in NLP. Our goal was to understand if
emotion-bearing texts allow to infer affective information that goes
beyond discrete emotion experiences as captured by emotion labels,
extending to their core motives of ``seeking more or less'' of a
stimulus.

Identifying coping strategies computationally could indeed benefit
emotion analysis in multiple ways, as it can be instrumental for
different downstream objectives, for instance to understand
argumentative strategies, revealing how people react to verbal
conflicts. Leveraging coping strategies is also valuable to establish
connections with other theoretical and computational frameworks. One
can identify coping strategies in a multi-task setup to better
identify the emotions of a text (e.g., based on Ekman's theory), or
explore their link with appraisal theories in event-centered emotion
analysis~\citep{Klinger2023a}, to explain how events (e.g., those
described in our scenarios) are evaluated or
re-appraised~\citep[cf.][]{li_herderich_goldenberg_2024,Morris2015}. Such
integrated approaches promise a comprehensive understanding of
emotions that is still ripe for investigation.

As a first contribution in this direction, we built a corpus for
coping detection via role-playing. Role-playing enabled us to overcome
the lack of available textual data annotated with coping
information. Moreover, it facilitated our attempt to communicate
abstract constructs to laypeople, and to gain control over many
variables (e.g., all people had the goal of realizing a coping
strategy, in response to the same controversial statements, on a small
set of topics) but at the same time it introduced some
artifacts. Notably, the expression and identification of coping
strongly relied on our ability to communicate them in $X$'s
descriptions, as well as people's ability to understand such
descriptions, to engage in role-playing, and to produce good-quality,
coping-expressing texts.

Therefore, both the quality of the texts and the classification
results are to be discussed while considering the challenges
introduced by our data collection procedure: the participants in our
study demonstrated a robust understanding of coping within a strict
experimental setup -- corroborating that data on this psychological
concept can be obtained via crowdsourcing; the texts they produced, on
the other hand, generally do not exhibit characteristics that allowed
our readers or automatic classifiers to perfectly discern the correct
coping strategies. It remains to be determined whether this is due to
the intentionality of coping expression (despite evidence that the
crowdworkers' acting was successful) or because these strategies are
naturally communicated in covert manners, making them hard to discern.
Overall, the emergence of coping from text appears less
straightforward than we initially hypothesized, but this conclusion is
not definitive, as one could still expect that, were the texts
produced for counter-arguing purposes (like ours) but spontaneously
(unlike ours), the attack, contact, distance, or reject mechanisms
might be clearer.  In fact, the possible recognition of coping, and
the idea that these strategies lend themselves for annotation is
corroborated by our analysis of their link to emotions -- with
meaningful patterns that find substantial correspondence in Roseman's
literature. This hints at the great potential of coping to be used for
future tasks.

\section*{Acknowledgements}
This work has been supported by the Deutsche Forschungsgesellschaft
(DFG, Project KL 2869/1-2, No.\ 380093645). It also received funding
from the Flemish Government under the Research Program Artificial
Intelligence (grant no. 174E07824) and under the Research Foundation -
Flanders (FWO, grant no. 1S96322N).  We further thank Kai Sassenberg
for his valuable input and Amit Goldenberg his insightful pointers
concerning negativity reframing and re-appraisal.

\section*{Ethical Statement}

The role-playing study has been approved by an ethical committee
collaborating with our affiliations. Crowdworkers were rewarded with a
fee that corresponds to 9£ \textit{per} hour for their participation.

Since the texts in \corpusname are based on conflictual dialogues,
they contain potentially sensitive material that might depict
real-life experiences of some individuals.  All crowdworkers were
warned beforehand, and were informed about the type of debate they
would have entered by taking part in the study. Their IDs in the
corpus have been encrypted to prevent identification.

Both the fictional interlocutor (whose utterances were produced by
ChatGPT) and $X$ (impersonated by people) express standpoints that can
have political significance.  We remain neutral with respect to both
sides.

\section*{Limitations}

While we deemed \corpusname appropriate to conduct a proof-of-concept
study, we acknowledge that its size is suboptimal to extrapolate
features of coping strategies, and train classification models.

Its construction relied on several preliminary rounds of crowdsourcing
which corroborated the difficulty of our tasks for laypeople. We
simplified it via role-playing, but this strategy required us to
embrace some simplifying assumptions. First and foremost, texts were
not generated while experiencing the emotion(s) necessary to implement
a coping strategy. The conflictual situations depicted in the
scenarios had the potential to stir an emotion, but we did not measure
their success in this sense, and we recognize that imagining to feel
an emotion is not the same (in terms of cognitive and physiological
processes) as actually feeling one.

To mitigate this problem, we assigned participants to the topics they
might feel more related to by opening the studies to crowdworkers with
specific demographics (e.g., women for the abortion topic). This
decision was also based on simplifying assumptions, namely, that
demographics approximate one's experiences and involvement in the
given topic, and that the understanding of laypeople would reflect in
their ability to generate texts.

Overall, our setup does not necessarily correspond to one that
psychological research on coping would adopt: albeit useful, coping
strategies still require us to reflect on how to more seamlessly
connect emotion-based NLP and the disciplines from which we borrow
theoretical constructs.

\bibliography{anthology}
\bibliographystyle{acl_natbib}

\clearpage

\appendix

\section{Describing Coping Strategies}
\label{sec:data-creation-descriptions}

The formulation of coping strategies described in
Section~\ref{ssec:describing-coping} was based on 4 crowdsourcing
rounds.  Round 1, 3, and 4 involved 100 participants each (25 per
coping strategy). In round 2, we only focused on attack and reject,
with the goal to maximize their difference (25 people participated for
each of these strategies).  We selected participants on the basis of
first and primary language (i.e., English), approval rate on past
Prolific studies (between 90\% and 100\%). Participants could only
take our study once (i.e., for no more than one coping strategy, in no
more than one round). The total cost of this phase was 210\pounds.

Each round comprised 4 questionnaires, one per coping strategy.  All
questionnaires included the same questions, one for the identification
of coping strategies and two others, to (1) rate how much of a given
list of emotions are likely felt by character $X$, when confronted
with an unpleasant situation, and (2) associate (the Big-Five)
personality traits to $X$. The emotion list included those analyzed
by~\citet{Roseman2013}: anger, contempt, disgust, dislike, distress,
fear, frustration, guilt, hope, joy, love, pride, regret, relief,
sadness, shame. Each of them was rated on a 4-item scale (i.e., from
{\it The emotion does not apply} to {\it The emotion very much
  applies}). We additionally collected personality traits because the
coping understanding task presents coping strategies in terms of
personalities.

Emotion information served to observe if the annotators' associations
of emotions to coping strategies mapped to the original taxonomy
proposed by Roseman, in which each strategy underlies a handful of
emotions (similar to what we did in the role-playing data, in
Section~\ref{ssec:writers}). Instead, information about the traits of
$X$ served to analyze if $X$'s descriptions led people to form similar
understandings of its persona, in terms of openness to experience,
conscientiousness, extroversion, agreeableness and emotional
stability.

We detail tasks and corresponding results below.

\subsection{Annotating Coping Descriptions}
\label{ssec:appendix-coping-description}

\begin{figure}
\centering
\begin{tabular}{c}
\textbf{Round 1} \\[.1cm]
\includegraphics[scale=.58]{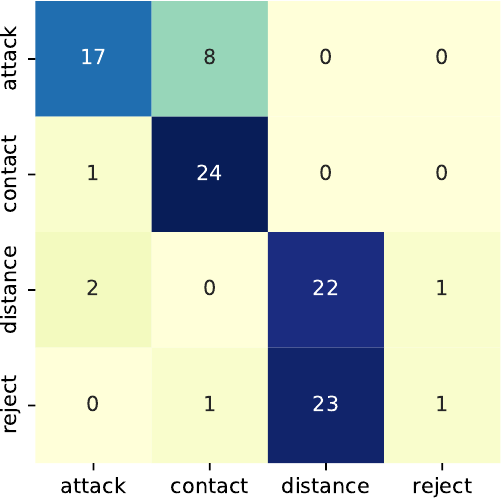}\\
\textbf{Round 2}\\[.1cm]
\includegraphics[scale=.58]{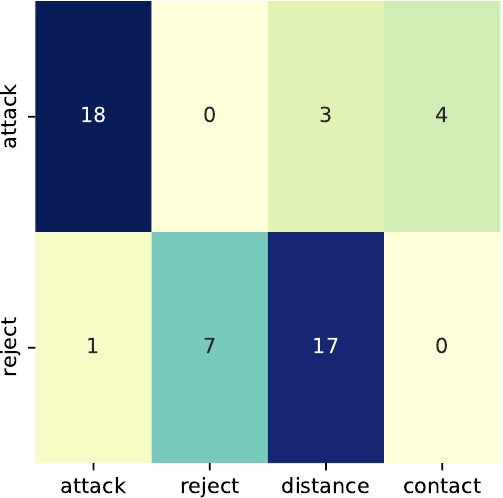}\\
\textbf{Round 3}\\[.1cm]
\includegraphics[scale=.58]{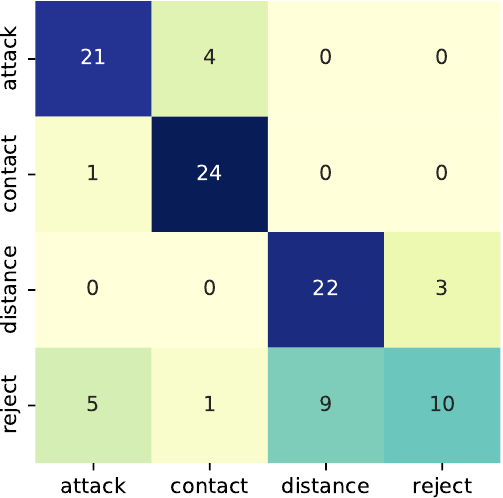}\\
\textbf{Round 4}\\[.1cm]
\includegraphics[scale=.58]{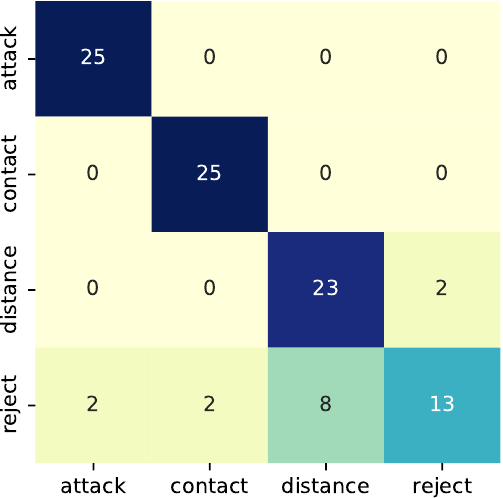}\\

\end{tabular}
\caption{Confusion matrices for the pre-test rounds. Rows are the
  annotated labels. Columns are original labels. Numbers correspond to
  the annotators (in total, 25 for each strategy).}
\label{fig:confusion-coping-pretest}
\end{figure}

The description of character $X$ in a coping-specific questionnaire
changed from round to round.
Figure~\ref{fig:confusion-coping-pretest} reports the confusion
matrices for the successive rounds of this crowdsourcing step,
indicating the number of annotators who correctly associated $X$
description to the corresponding coping strategy (diagonal cells) and
how many made incorrect associations (off-diagonal numbers).

The descriptions entering the role-playing guidelines are those found
in the last round (see
Table~\ref{tab:pre-study-descriptions-4}). Thanks to such
descriptions, the annotators reached perfect understanding of attack
and contact, 92\% of the annotators correctly chose the distance
strategy, and 52\% made the correct inference for reject.

\begin{table*}
\centering\small
\begin{tabular}{p{12cm}p{2cm}}
\toprule
\multicolumn{2}{c}{\textbf{Strategy: Attack}}\\
\cmidrule(r){1-2}
\textbf{Description}: X is a very opinionated person, who comes across as aggressive and combative.
When problems or unpleasant situations arise, X responds with a hardened resolve. X gives their all to overcome obstacles, and never shies away from a conflict. In fact, this character faces negative situations fiercely and with intense energy, taking action personally in order to right the wrong. For this reason, one can often see X all tense and pumped up.&\multirow{2}{*}{\includegraphics[scale=.25]{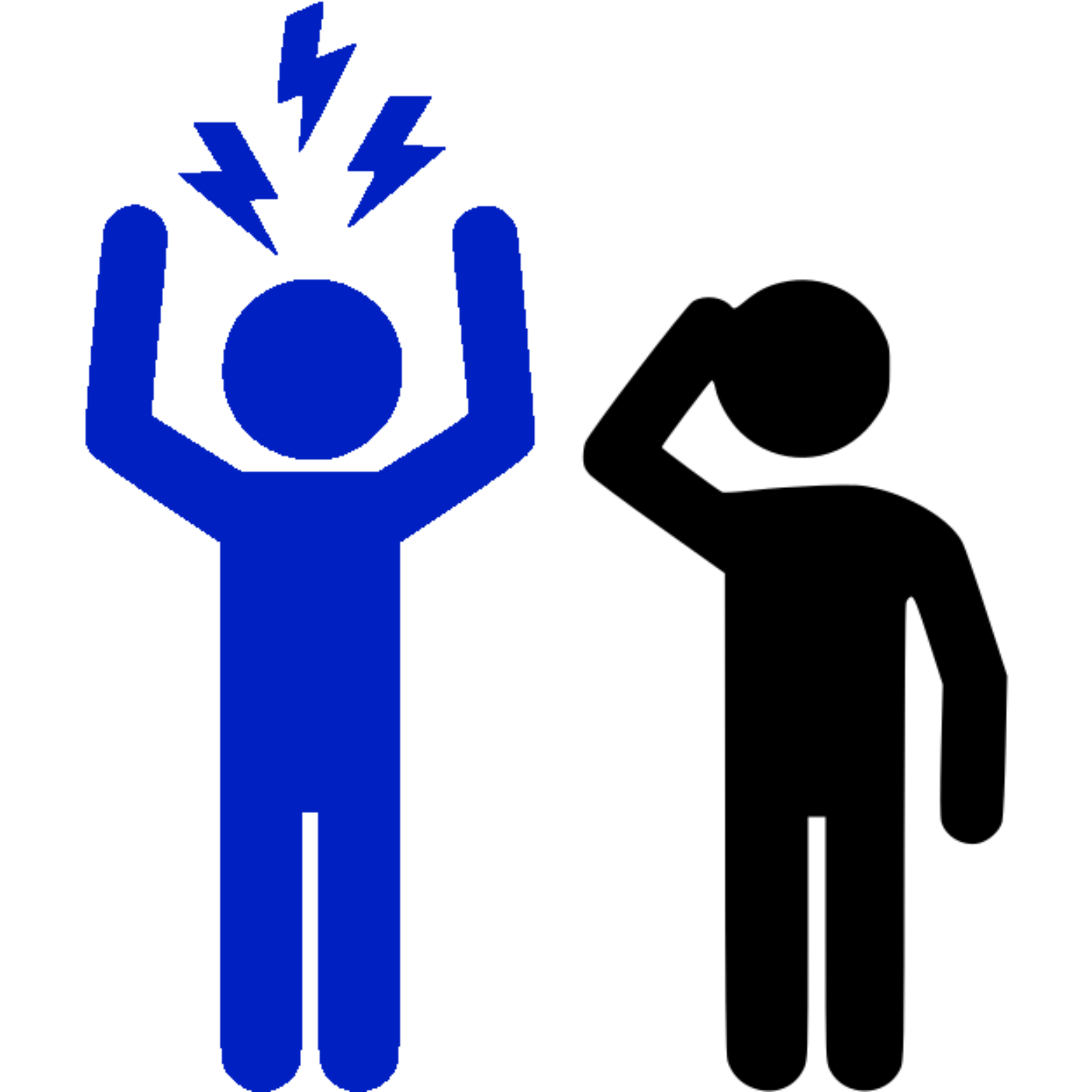}}\\

\textbf{Short summary}: X rights the wrong in an aggressive manner, like in the picture below (X is shown in blue). &\\\\

\midrule
\multicolumn{2}{c}{\textbf{Strategy: Contact}}\\
\cmidrule(r){1-2}
{\bf Description}: This character comes across as a calm, understanding, and very approachable person. For X, communication serves to unite people. It is an opportunity to exchange opinions, acknowledging the diversity of perspectives among individuals. When problems or unpleasant situations arise, this character responds with a constructive attitude. X expresses ideas with confidence, trying to solve problems in a respectful manner. X can effectively engage in discussions also with people having contrasting opinions.&\multirow{2}{*}{\includegraphics[scale=.25]{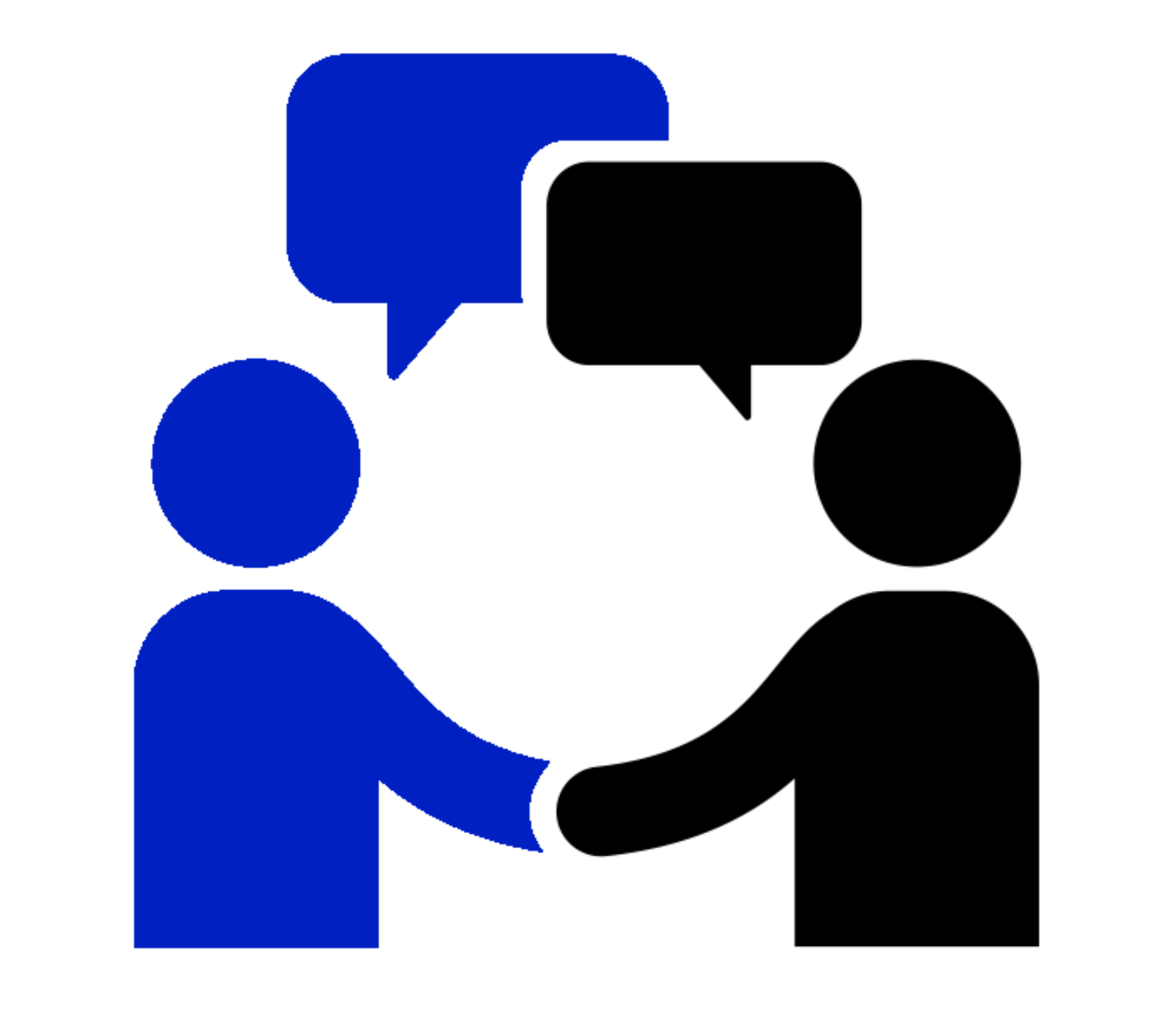}}\\

\textbf{Short summary}:  X rights the wrong with an open mindset, like in the picture below (X is shown in blue).&\\\\\\[.5pt]

\midrule
\multicolumn{2}{c}{\textbf{Strategy: Distance}}\\
\cmidrule(r){1-2}
\textbf{Description}: In life, X has managed to create a peaceful mental space as a protection from the outside world. This character always seeks emotional tranquillity, and finds solace in self-reflection. When problems or unpleasant situations arise, X feels powerless, not too capable to change the external world. So here's X's defense strategy against all negativity: staying away from stressful things, events, people or thoughts, to minimize their influence. At the cost of coming across as an avoidant person, X does not engage in conflicts and uncomfortable situations. & \multirow{2}{*}{\includegraphics[scale=.25]{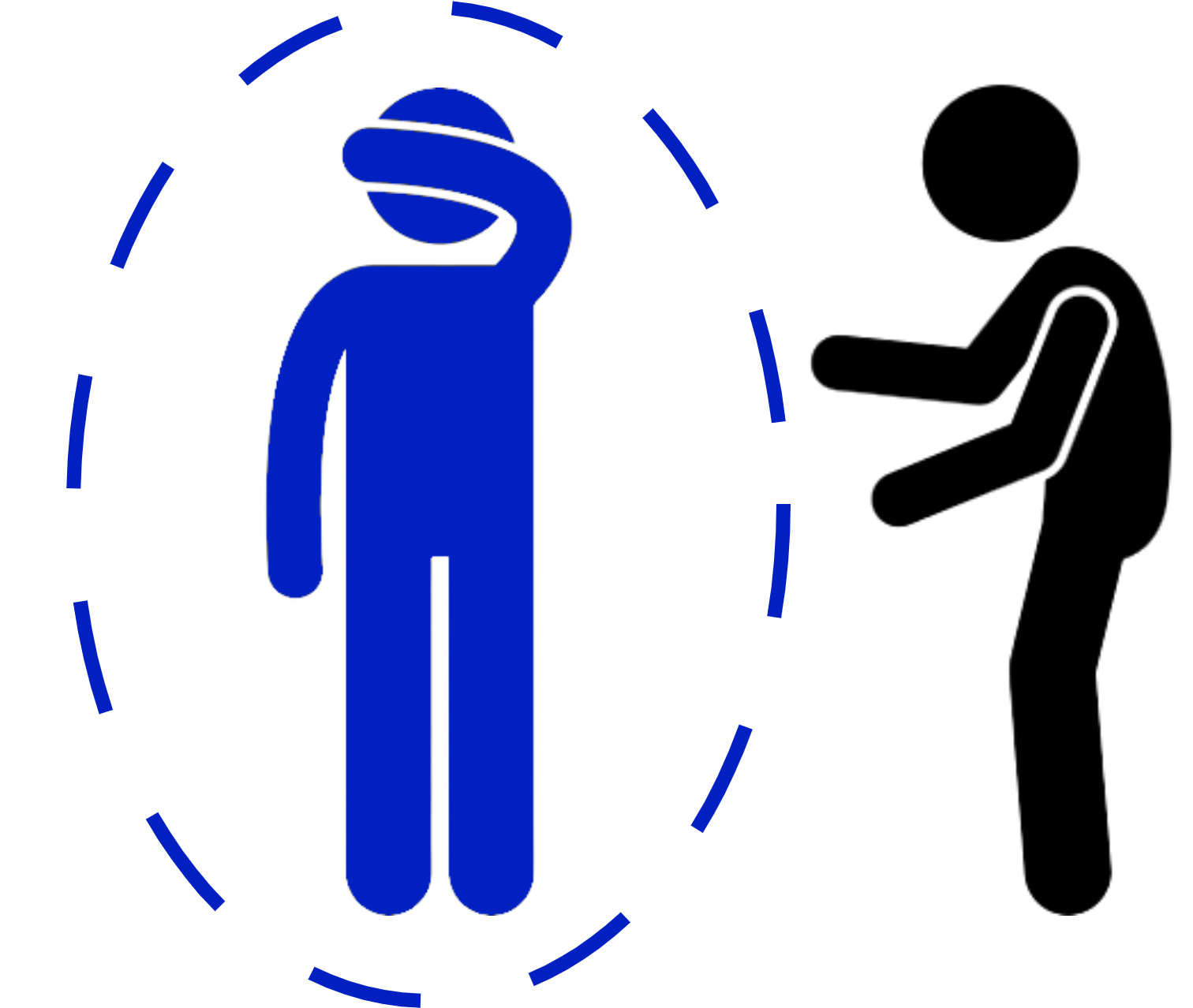}}\\

\textbf{Short summary}: X seeks shelter from unpleasant situations in a quiet inner space, like in the picture below
(X is shown in blue).&\\\\[.5pt]

\midrule
\multicolumn{2}{c}{\textbf{Strategy: Reject}}\\
\cmidrule(r){1-2}

\textbf{Description}: This person has a judgmental attitude towards others. X easily disagrees with people, and is often displeased with their behaviors and beliefs. Here's X’s motto in life: do not let anybody interfere with your well-being. This character keeps negativity away at all costs. It's not that X hides from problems and tries to find solace in a peaceful mental space, it's more of a resolute attempt to refuse unpleasant situations. It's also not a lack of courage; in fact, X expresses opinions with confidence, and precisely by this, X manages to dismiss conflicts with other people. & \multirow{2}{*}{\includegraphics[scale=.25]{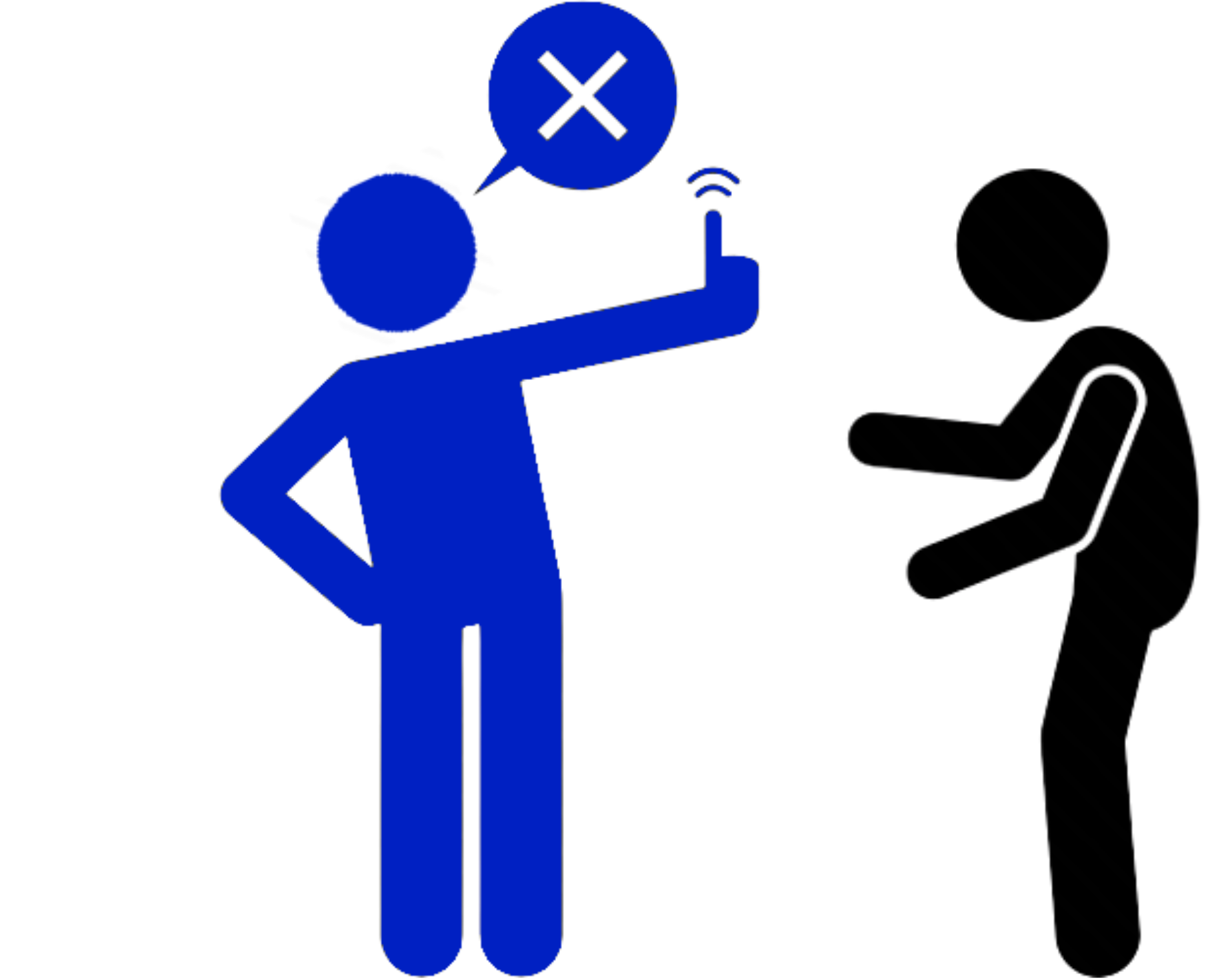}}\\

\textbf{Short summary}: 
X's approach to life is a firm refusal of negativity, like in the picture below
(X is shown in blue). &
\\\\

\bottomrule
\end{tabular}
\caption{Descriptions of $X$ used in the final round of coping recognition and in the role-playing crowdsourcing phase. Each copying strategy is presented as a description of character $X$, and in a more concise version (short summary), and with the reported picture.}
\label{tab:pre-study-descriptions-4}
\end{table*}

\subsection{Additional Tasks}
Besides associating character's $X$ description to a one-liner summary of the corresponding coping strategy, the annotators judged emotions and personality traits.

\begin{figure*}
\centering
\begin{tabular}{cccc}
&& \textbf{Correct Annotators} & \textbf{Wrong Annotators}\\
\raisebox{2.3cm}{\rotatebox{90}{\textbf{Attack}}} && \includegraphics[scale=.38]{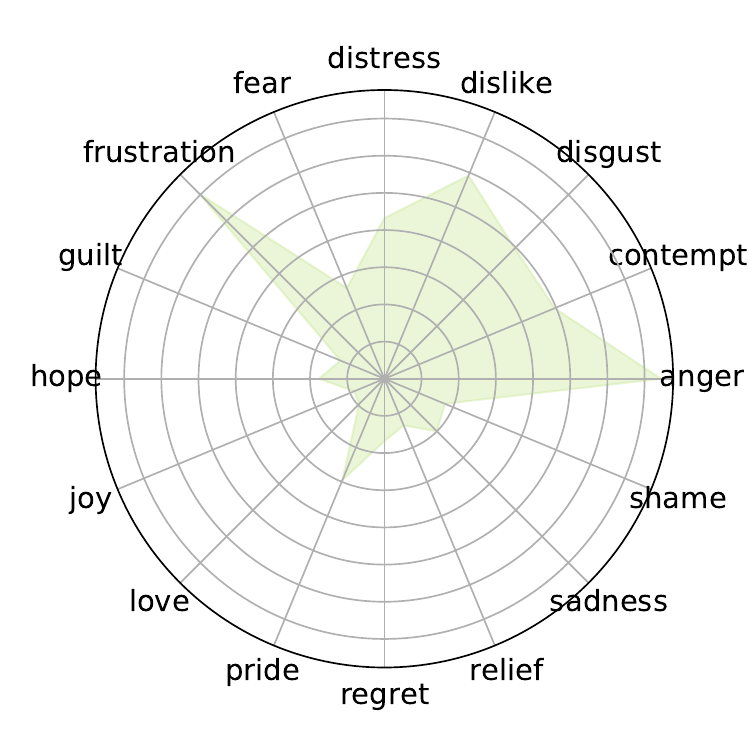}
&\\[.2cm]
\raisebox{2.3cm}{\rotatebox{90}{\textbf{Contact}}} && \includegraphics[scale=.38]{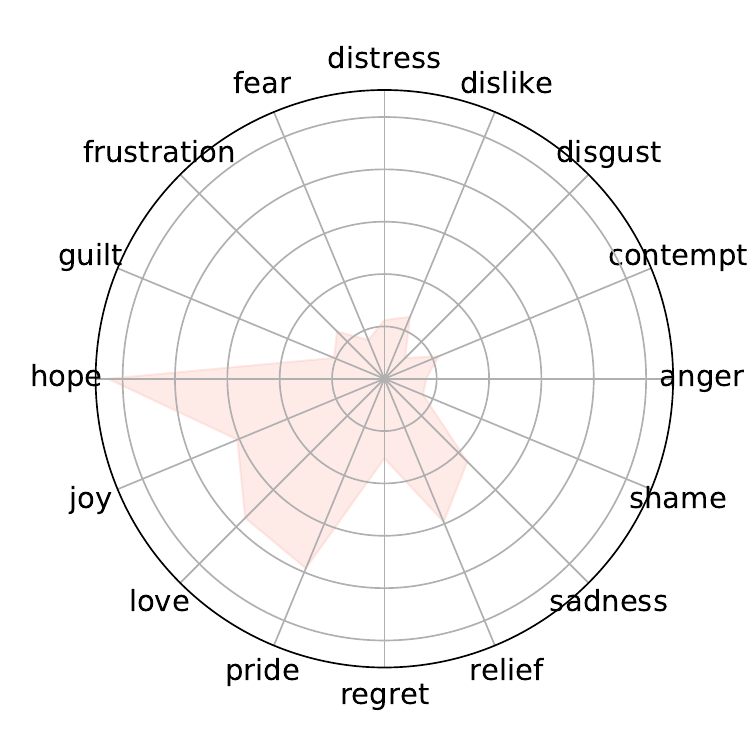}
& 
\\[.2cm]
\raisebox{2.3cm}{\rotatebox{90}{\textbf{Distance}}} && \includegraphics[scale=.38]{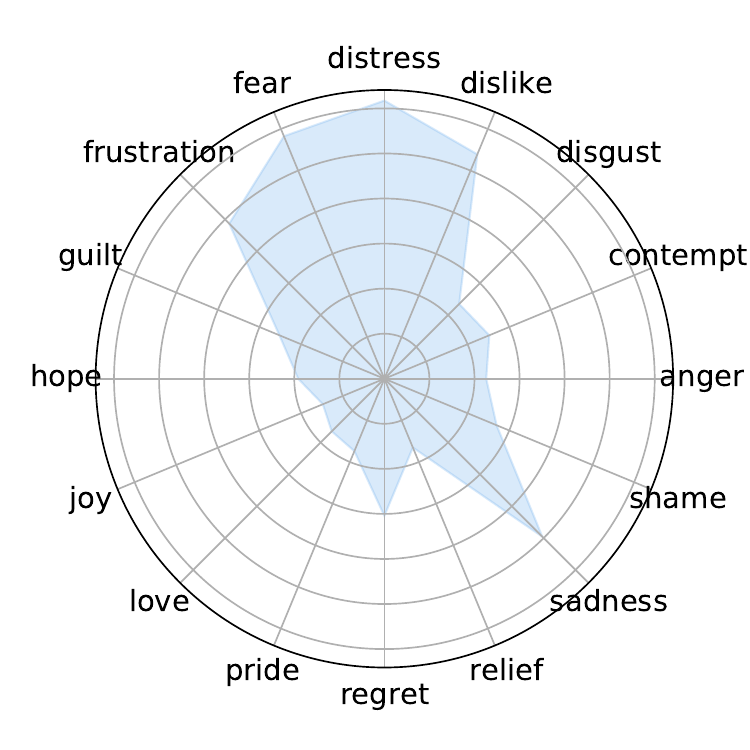}
& 
\includegraphics[scale=.38]{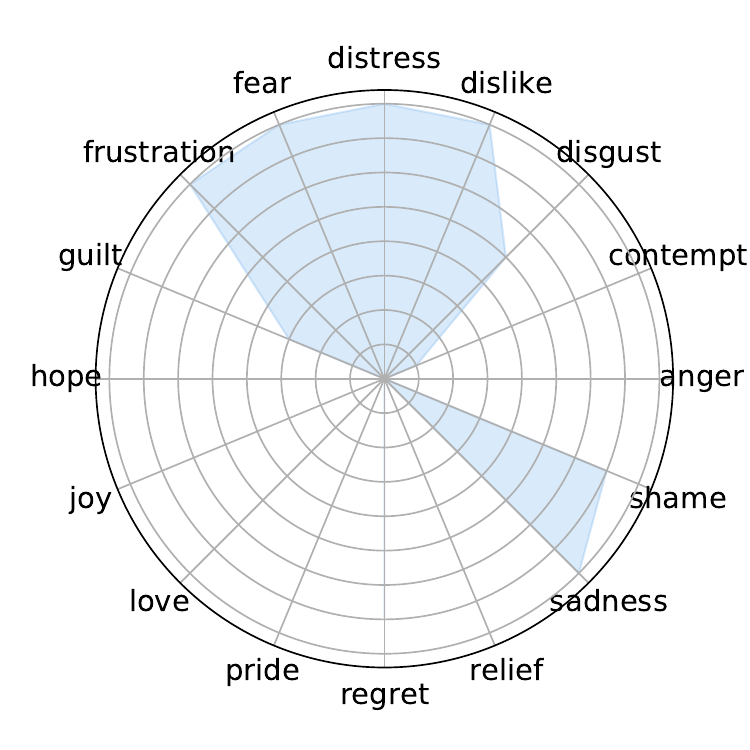}\\[.2cm]
\raisebox{2.3cm}{\rotatebox{90}{\textbf{Reject}}} && \includegraphics[scale=.38]{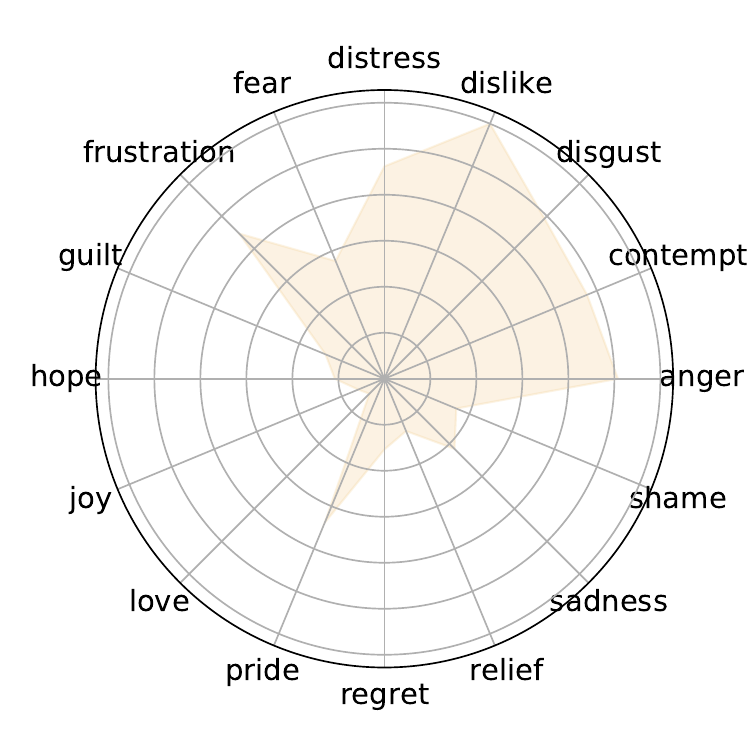}
& 
\includegraphics[scale=.38]{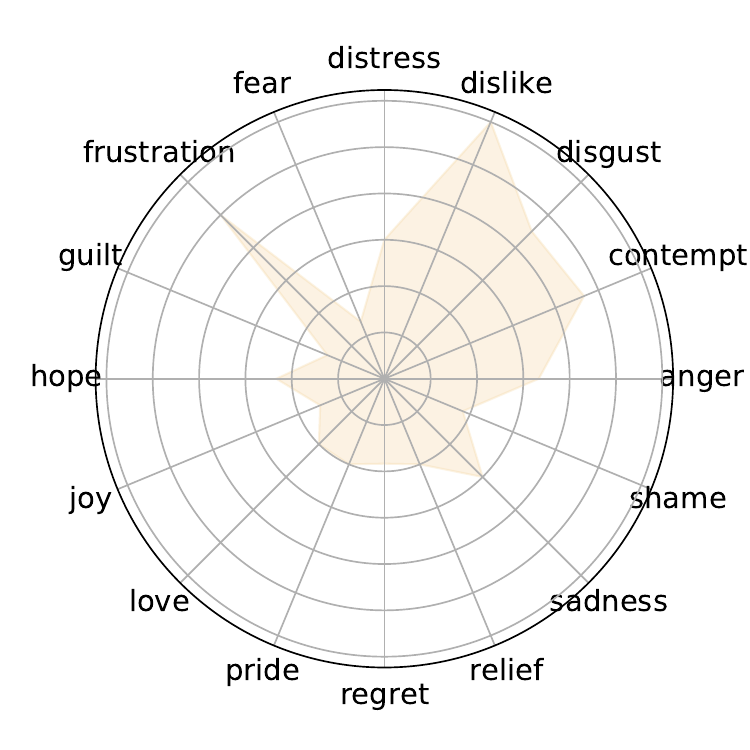}\\[.2cm]
\end{tabular}
\caption{Plots of the emotion answers (averages) for the last crowdsourcing round of coping strategy recognition, divided by annotators who were right about the coping strategy (Correct Annotators) and those who did not correctly recover the strategy from the description (Wrong Annotators). For \textit{attack} and \textit{contact}, all crowdworkers were correct.}
\label{fig:spiders-emotions-pretest-round3}
\end{figure*}

\paragraph{Emotions.}
In order to better assess the performance the annotators, we investigated how similar the emotion understandings of
wrong and correct annotators are. More precisely, we observe:
\begin{enumerate}
    \item if the annotators rated the emotions similarly, despite
    their coping choices being different (if so, we can move on to the generation phase);
    \item to what extent they captured the
    ``correct'' emotions (i.e., those assigned by Roseman to each coping strategy) -- once more, this would be a sign that we can proceed with role-playing phase of crowdsourcing.
\end{enumerate}
The answer to point 1. emerges from
Figure~\ref{fig:spiders-emotions-pretest-round3},
Table~\ref{attack-contact-emo-pretest-round4} and Table~\ref{distance-reject-emo-pretest-round4},
which contain results from the last round.

Figure~\ref{fig:spiders-emotions-pretest-round3} shows the average
values assigned to each emotion, for each coping strategy.  The values
obtained from the annotators who correctly understood the coping
strategy are side-by-side the values computed on the answers of the
annotators who did not get the correct strategy.

While the figure tells us ``how much'' the annotators rated an
emotion, Table~\ref{attack-contact-emo-pretest-round4} and
~\ref{distance-reject-emo-pretest-round4} summarize this information
in a binary way (did the annotators assign a given emotion to a coping
strategy?).  The symbol \ding{51} means that a given emotion was
recognized.  Note that emotions were rated on a scale from $1$ to $5$.
To decide if an emotion counts as correctly recognized, we observe if
it has an average rating of $>$2.  The numbers next to \ding{51}
correspond to the ranking of the emotion (e.g., $1$ means that the
emotion for that coping strategy had the highest average rating).

From both Figure~\ref{fig:spiders-emotions-pretest-round3}
and the tables, we see that the emotions captured by the two groups of annotators overlap for all coping strategies.
\bigskip

Concerning point 2., we answer it by looking at
Table~\ref{attack-contact-emo-pretest-round4} and
Table~\ref{distance-reject-emo-pretest-round4}. Both report the
annotators' judgments next to the correct emotions indicated by
Roseman.  There, we see that for none of the coping strategies the
annotators understood all relevant emotions. However, the correct
emotions that the crowdworkers captured have the highest ranks (e.g.,
focusing on attack, anger and frustration have the highest ratings,
for both groups of annotators).

\begin{table*}
\centering\small
\renewcommand{\arraystretch}{.8}
\begin{tabular}{p{2.6cm}p{1.8cm}p{2.6cm}p{1.8cm}p{2.6cm}}
\toprule
 & \multicolumn{2}{c}{\textbf{Attack}} & \multicolumn{2}{c}{\textbf{Contact}}\\
 \cmidrule(r){2-3} \cmidrule(r){4-5}
\textbf{Emotions} & Original & Correct Annotat. & Original & Correct Annotat.\\
\cmidrule(r){2-2}\cmidrule(r){3-3}\cmidrule(r){4-4}\cmidrule(r){5-5}
Anger & \ding{51} & \ding{51}, 1 & &\\
Contempt &  & \ding{51}, 5 & &\\
Disgust &  & \ding{51}, 4& &\\
Dislike &  & \ding{51}, 3& &\\
Distress &  & \ding{51}, 6 & &\\
Fear &  & &&\\
Frustration & \ding{51} & \ding{51}, 2&&\\
Guilt & \ding{51} & &&\\
Hope&  & &\ding{51}&\ding{51}, 1\\
Joy&  & &\ding{51}&\\
Love&  & &\ding{51}&\\
Pride&  & &\ding{51}&\\
Regret&  & &&\\
Relief&  & &\ding{51}&\\
Sadness&  & &&\\
Shame &  & &&\\
Surprise &  & &&\\
\bottomrule
\end{tabular}
\caption{Emotions for the strategy Attack and Contact as indicated by~\citet{Roseman2013} (column Original) and understood by our annotators in the last round of the coping annotation phase of crowdsourcing. All annotators in this round identified the correct coping strategy (Correct Annotators). \ding{51} means that an emotion average rating was $>2$. Numbers are rankings of emotion averages.}
\label{attack-contact-emo-pretest-round4}
\end{table*}

\begin{table*}
\centering\small
\renewcommand{\arraystretch}{.8}
\begin{tabular}{p{2.6cm}p{1.8cm}p{1.8cm}p{1.8cm}p{1.8cm}p{1.8cm}p{1.8cm}}
\toprule
 & \multicolumn{3}{c}{\textbf{Distance}} & \multicolumn{3}{c}{\textbf{Reject}}\\
 \cmidrule(r){2-4} \cmidrule(r){5-7}
\textbf{Emotions} & Original & \makecell{Correct\\Annotat.} & \makecell{Wrong\\Annotat.}&Original & \makecell{Correct\\Annotat.} & \makecell{Wrong\\Annotat.}\\
\cmidrule(r){2-2}\cmidrule(r){3-3}\cmidrule(r){4-4}
\cmidrule(r){5-5}\cmidrule(r){6-6}\cmidrule(r){7-7}
Anger &  & & &  & \ding{51}, 2& \\
Contempt &  & & & \ding{51} & \ding{51}, 4& \ding{51}, 3\\
Disgust &  & & &  \ding{51}& \ding{51}, 3& \ding{51}, 4\\
Dislike & \ding{51} & \ding{51}, 3& \ding{51}, 1&  & \ding{51}, 1 & \ding{51}, 1\\
Distress & \ding{51} & \ding{51}, 1& \ding{51}, 2&  & \ding{51}, 5& \\
Fear & \ding{51} &\ding{51}, 2 & \ding{51}, 3&  & & \\
Frustration &  &\ding{51}, 5 & \ding{51}, 4&  & \ding{51},  6& \ding{51}, 2\\
Guilt &  & &&  & & \\
Hope&  & &&  & & \\
Joy&  & &&  & & \\
Love&  & &&  & & \\
Pride&  & &&  & & \\
Regret& \ding{51} & & \ding{51}, 6&  & & \\
Relief&  & &&  & & \\
Sadness& \ding{51} & \ding{51}, 4& \ding{51}, 5&  & & \\
Shame &  && \ding{51}, 7&\ding{51}  & & \\
Surprise&  & &&  & & \\
\bottomrule
\end{tabular}
\caption{Emotions for the strategy Distance and Reject as indicated by~\citet{Roseman2013} (column Original) and understood by our annotators in the last round of the coping annotation phase of crowdsourcing. Annotators are divided into those who identified the correct coping strategy (column Correct Annot.) and those who failed the task (column Wrong Annot.). \ding{51} means that an emotion average rating was $>2$. Numbers are rankings of emotion averages.}
\label{distance-reject-emo-pretest-round4}
\end{table*}

\paragraph{Personality Traits.}
Crowdworkers estimated how much the 10 items in the questionnaire
of~\citet{gosling2003very} might apply to character $X$.  The
questionnaire evaluated openness to experience by contrasting ``open
to new experiences and complex'' against ``conventional and
uncreative''; conscientiousness through the trait ``dependable and
self-disciplined'' versus ``disorganized and careless''; extraversion
by comparing ``extraverted and enthusiastic'' with ``reserved and
quiet''; agreeableness by distinguishing between ``sympathetic and
warm'' and ``critical and quarrelsome''; and emotional stability by
assessing ``calm and emotionally stable'' against ``anxious and easily
upset''. Respondents rated $X$ on these dichotomous descriptors using
a 7-point Likert scale ranging from ``disagree strongly'' to ``agree
strongly''.

\clearpage
\section{Generating Scenarios}
\label{sec:appendix-coping-scenario}

We identified the topics via ChatGPT-4, asked to list 20 sensitive
topics for specific categories of people.  This step was independently
performed by three researchers, who later integrated and refined their
outputs in a set of \textit{5} topics: abortion, immigration, racism,
LGBTQ+, drugs.

For the scenario generation, ChatGPT-4 had to refrain from
characterizing the gender of the interlocutors (hence, the names $X$
and $Y$). Scenarios were produced between August 14 and October 31
(2023) with ChatGPT-4.

Table~\ref{tab:scenarios-chat} shows how we prompted ChatGPT-4 (we
show it for only one topic and one scenario for brevity) and
Table~\ref{tab:scenarios-per-topic} reports all scenarios we used in
our study.

\onecolumn
{\small
\begin{longtable}{p{15cm}}
\caption{ChatGPT-4 generation of scenarios. Green rows correspond to humans, white rows to the assistant's replies. For simplicity, we only report one generated scenario.}\label{tab:scenarios-chat} \\
\toprule
\rowcolor{lightgreen}
I am conducting a study about Roseman's coping strategies (i.e., attack, contact, distance and reject). My goal is to study how each strategy realises in language when speakers discuss about controversial topics. Imagine that during an interaction, one of the interlocutors expresses an opinion that makes another uncomfortable; the latter thus responds with a verbal utterance and a non-verbal reaction while being "under the influence" of a coping strategy.\\

Certainly! Studying how Roseman's coping strategies manifest in language, especially during discussions on controversial topics, is an intriguing area of inquiry. When individuals discuss controversial topics and face conflicting or uncomfortable opinions, they might adopt one or more of Roseman's coping strategies in their verbal and non-verbal behavior.

Here's a breakdown of each strategy along with potential verbal and non-verbal manifestations:\\
1.  {\bf Attack}:

{\bf Verbal}: An individual might use aggressive language, accusations, or challenges. They might say things like, "That's completely wrong," "How can you even believe that?", or "You clearly haven't done your research." 

{\bf Non-Verbal}: Physically leaning forward, intense eye contact, clenched fists, raised voice, or even pointing fingers. The individual's face might show signs of anger or frustration.\\

2.  {\bf Contact}:
    
   {\bf Verbal}: The individual would seek clarification, ask questions, or try to understand the other's viewpoint. They might say, "Can you explain why you feel that way?", "I'm curious to know more about your perspective," or "Let's find some common ground."
    
    {\bf Non-Verbal}: Leaning in, maintaining steady but non-threatening eye contact, nodding to indicate active listening, and open body posture. The face might reflect genuine curiosity or concern.\\

3.  {\bf DISTANCE}:
    
     {\bf Verbal}: An individual might divert the topic, use vague language, or minimize their involvement in the conversation. They might say, "Let's discuss this later," "I need some time to think about it," or simply change the topic altogether.
     
   {\bf Non-Verbal}: Looking away, physically leaning back, creating physical distance, crossing arms as a defensive posture, or checking one's phone/watch. The face might show signs of discomfort or avoidance.\\

4.  {\bf Reject}:
    
    {\bf Verbal}: The individual might dismiss the topic or the person's opinion outright, often without offering a counter-argument. They might say, "That's not even worth discussing," "I can't entertain that thought," or "We'll never agree on this, so let's drop it."
    
    {\bf Non-Verbal}: Eye-rolling, shaking the head, waving one's hand dismissively, or even walking away. The face might display contempt, disdain, or indifference.\\

For a comprehensive study, you may also want to consider the tone of voice, the specific linguistic constructions used, the rhythm and pace of speech, and other nuanced non-verbal cues like pupil dilation or micro-expressions. Moreover, always remember that cultural and individual variations can affect how these coping strategies manifest. It would be beneficial to take these variations into account when analyzing and interpreting your findings.\\\\

\rowcolor{lightgreen}
With my next prompts, I'm going to ask you to conduct the following task. 

{\bf Task} Depict three scenarios (each set in a different social context, briefly described) in which two characters, X and Y, interact. Y produces a controversial statement about a given topic that I will specify. That opinion could elicit an emotional reaction in X. Each scenario must be structured like so: 

- SCENARIO AND CONTROVERSIAL OPINION 

- VERBALIZED REACTION AND NON VERBALIZED REACTION ASSUMING THAT PERSON B ADOPTS THE COPING STRATEGY OF ATTACK 

- VERBALIZED REACTION AND NON VERBALIZED REACTION ASSUMING THAT PERSON B ADOPTS THE COPING STRATEGY OF CONTACT 

- VERBALIZED REACTION AND NON VERBALIZED REACTION ASSUMING THAT PERSON B ADOPTS THE COPING STRATEGY OF DISTANCE - VERBALIZED REACTION AND NON VERBALIZED REACTION ASSUMING THAT PERSON B ADOPTS THE COPING STRATEGY OF REJECT 

Requirements: 

- SCENARIO AND CONTROVERSIAL OPINIONS should span maximum 6 sentences; 

- for each scenario, the controversial statement must focus on a different aspect of the issue; 

- X must have a verbalized and non-verbalized reaction corresponding to one coping strategy; 

- each scenario should feature a different power relationship between X and Y, with them being peers in the first scenario (none holds more power than the other), X being the dominant individual in the second, and Y being the dominant individual in the third; 

- when describing the scenario, do not mention the demographics of the interlocutors, but specify their role to one another (e.g., who is Y to X?) as to make the reader understand their relationship.

I will later input the topics and you will generate the scenarios. For now, just confirm if the task is clear.\\\\

The task is clear. Once you provide the topics for the controversial statements, I will generate scenarios based on the structure and requirements you've outlined. Just to recap:

1.  We'll have three scenarios set in distinct social contexts.

2.  The nature of the relationship between X and Y will differ in each scenario: peers in the first, X dominant in the second, and Y dominant in the third.

3.  Y will make a controversial statement in each scenario.

4.  X's reaction will be provided for each of the four coping strategies, both verbally and non-verbally.

5.  The demographic details of the interlocutors won't be specified, but their roles/relationship will be.

Please provide the topics for the controversial statements, and I'll proceed with creating the scenarios.\\\\

\rowcolor{lightgreen}
Please generate three different scenarios for the topic immigration policy and border control (at least one of them should focus on islamophobia).\\

{\bf SCENARIO 1: Colleagues at a Company Coffee Break}

{\bf SCENARIO AND CONTROVERSIAL OPINION}  
X and Y are colleagues who work in the same department and have known each other for a year. During a coffee break, they discuss current events. Y, stirring their coffee, remarks, "Honestly, I think our country's immigration policy is too lenient. We're just letting anyone in, and it's bound to create chaos."

\begin{itemize}
\item {\bf ATTACK}

{\bf Verbalised Reaction} "That's such a narrow-minded view, Y! It's people like you who perpetuate unnecessary fear." 
 
{\bf Non-Verbalised Reaction} X's eyes narrow, lips tighten, leans forward aggressively, fists clenching the edge of the table.

\item{\bf CONTACT}

{\bf Verbalised Reaction} "Why do you feel that way, Y? Do you think a stricter policy would really solve the perceived problems?"

{\bf Non-Verbalised Reaction} X tilts head slightly, maintains steady eye contact, eyebrows raised in a questioning manner.

\item{\bf DISTANCE}

{\bf Verbalised Reaction} "Hmm, interesting perspective. Have you tried the new blend of coffee they have here?"

{\bf Non-Verbalised Reaction} X takes a long sip from their mug, looks away briefly, and shifts weight from one foot to the other.

\item{\bf REJECT}

{\bf Verbalised Reaction} "I don't think this is the place or time to discuss such matters. Let's get back to work."

{\bf Non-Verbalised Reaction} X sighs, rolls eyes discreetly, and starts gathering their things.
\end{itemize}

\\
\bottomrule
\end{longtable}
}

\onecolumn
\begingroup
  \small
  \begin{tabularx}{0.98\linewidth}{p{3cm}X}
    \toprule
  \multicolumn{2}{c}{\textbf{Immigration}}\\
\midrule
\textbf{X-Y Relationship} & X and Y are colleagues who work in the same department and have known each other for a year.\\
\textbf{Scenario 1} & During a coffee break, they discuss current events. Y, stirring its coffee, remarks, ``Honestly, I think our country's immigration policy is too lenient. We're just letting anyone in, and it's bound to create chaos.''\\
\cmidrule(r){1-1}\cmidrule(r){2-2}
\textbf{X-Y Relationship} & Y is a student who stays back after class to discuss an assignment with X, the teacher.\\
\textbf{Scenario 2} & During their conversation, Y says, ``It's obvious that the recent surge in Muslim immigrants is just a strategy for them to dominate our culture.''\\
\cmidrule(r){1-1}\cmidrule(r){2-2}
\textbf{X-Y Relationship} & X is an applicant for a managerial role, while Y is the CEO of the company conducting the interview.\\
\textbf{Scenario 3} & As they discuss potential challenges, Y mentions, ``We've had issues with foreign workers not blending in. Honestly, open borders are more of a problem than a solution.''\\
\midrule
\multicolumn{2}{c}{\textbf{Racism}}\\
\midrule
\textbf{X-Y Relationship} & X is a classmate to Y\\
\textbf{Scenario 1} & During a university class discussion on historical racial events, B confidently states, ``People keep talking about systemic racism, but I believe that's just an excuse for those who don't want to work hard. If you look around, everyone has the same opportunities today.''\\
\cmidrule(r){1-1}\cmidrule(r){2-2}
\textbf{X-Y Relationship} & X is an instructor, and Y is a regular attendee of a fitness class.\\
\textbf{Scenario 2} & During a water break, Y comments to a friend, but loud enough for X to hear, ``I still can't believe they hired a Latino as our instructor.''\\
\cmidrule(r){1-1}\cmidrule(r){2-2}
\textbf{X-Y Relationship} & X and Y are siblings, having dinner at their parents' home.\\
\textbf{Scenario 3} & As they discuss friends and relationships, Y says, ``You know I've always believed people should stick to their own kind. It's just easier.''\\
\midrule
\multicolumn{2}{c}{\textbf{Drugs}}\\
\midrule
\textbf{X-Y Relationship} & X is a concerned parent, and Y is a local politician.\\
\textbf{Scenario 1} & During a town hall meeting, Y proposes, ``Legalizing all drugs would cut down on crime and help manage addiction better through official channels.''\\
\cmidrule(r){1-1}\cmidrule(r){2-2}
\textbf{X-Y Relationship} & X is a podcast host, and Y is an expert in public health.\\
\textbf{Scenario 2} & During the recording of a podcast discussion, Y mentions, ``The 'War on Drugs' has largely been ineffective. A harm reduction approach is more viable.''\\
\cmidrule(r){1-1}\cmidrule(r){2-2}
\textbf{X-Y Relationship} & X and Y are friends.\\
\textbf{Scenario 3} & X and Y are friends catching up over coffee. As the topic shifts, Y shares, ``I've been reading about how some countries are decriminalizing drugs, and it seems like a good move. It focuses on treating users instead of punishing them.''\\
\midrule
\multicolumn{2}{c}{\textbf{Abortion}}\\
\midrule
\textbf{X-Y Relationship} & X and Y are both volunteers at a local food drive.\\
\textbf{Scenario 1} & While sorting through donations, Y mentions, ``Honestly, limiting access to abortion is just another way of controlling women. Everyone should have a say about their own bodies.''\\
\cmidrule(r){1-1}\cmidrule(r){2-2}
\textbf{X-Y Relationship} & X is an attendee at a charity gala, and Y is the guest speaker.\\
\textbf{Scenario 2:} & In Y's speech about women's health, they assert, ``It's time to prioritize the rights of women by ensuring unrestricted access to abortion services.''\\
\cmidrule(r){1-1}\cmidrule(r){2-2}
\textbf{X-Y Relationship} & At a charity gala dinner, X is a guest, and Y is the keynote speaker.\\
\textbf{Scenario 3} & During the speech, Y mentions, ``In a world where we're fighting for so many rights, I wish more emphasis was placed on the rights of the unborn.''\\
\midrule
\multicolumn{2}{c}{\textbf{LGBTQ+}}\\
\midrule
\textbf{X-Y Relationship} & X and Y are siblings, sitting together at a large family gathering.\\
\textbf{Scenario 1} & While discussing recent events, Y comments, ``I don't get this whole LGBTQ+ thing. There are just two genders. Why complicate things?''\\
\cmidrule(r){1-1}\cmidrule(r){2-2}
\textbf{X-Y Relationship} & X is a department manager, and Y is a colleague from a different department.\\
\textbf{Scenario 2} & While having coffee, Y says, ``These workplace LGBTQ+ training sessions are so unnecessary. Why can't people just keep their personal lives out of work?''\\
\cmidrule(r){1-1}\cmidrule(r){2-2}
\textbf{X-Y Relationship} & X is the class president, and Y is an old classmate.\\
    \textbf{Scenario 3} & During the reunion, Y whispers to X, ``Have you noticed? There are so many gay couples here now. It wasn't like this during our school days.''\\
    \bottomrule
\end{tabularx}
\captionof{table}{Scenarios used in the role-playing phase of data creation, divided by topic.\label{tab:scenarios-per-topic}}
\endgroup
\twocolumn

\clearpage

\onecolumn
\section{Role-Playing}
\label{sec:appendix-role-play}

While we assume that the participants can simulate coping strategies
and propagate them in text, doing that with laypeople in an artificial
setup is challenging. We therefore conducted the task in different
rounds, to experiment with different implementations of the
role-playing task.  In the first round, we assigned participants to
one $X$ to impersonnate.  Observing that this setup complicated the
role-playing performance, in the following rounds we assigned
participants to the $X$ that they mapped to the right $s$ in $t$: we
showed them the 4 $X$'s descriptions and allowed them to proceed with
the subsequent questions based on the $X$ that they label correctly.

Concerning the participants, we opened the study to crowdworkers with
a Prolific approval rate between 90\% and 100\%. We detail the
requirements for participation in Table~\ref{tab:participants},
divided by topic (the cross symbol in the table means we did not use a
given filter).  At each round, we excluded participants from previous
rounds.  For most studies, we selected participants with English as
their first language, and coming from the following countries: UK, US,
Canada, Ireland, New Zealand, Australia. The asterisk means we use
only a subset.

We reject participants who did not write in first person, did not use direct speech,
who produced ungrammatical or too short text, did not disagree with $Y$, or used
too little time to answer (less than 3 minutes).\\[1cm]

\begingroup
\centering\small
\begin{tabular}{p{3cm}p{10cm}}
\toprule
\textbf{Topic} & \textbf{Allowed Participants}\\
\cmidrule(r){1-1}\cmidrule(r){2-2}
Immigration$\dagger$ & Immigrated to the country where currently living. Place where spent most time before turning 18: Poland, Germany, India, Pakistan, Ireland, Romania, Mexico, China, Philippines, Dominican Republic, Cuba (i.e., the countries where majority of immigrants in UK and US come from).\\
Racism & Black, Asian, Mixed, Other\\
Drugs  & Have taken part in alcohol therapy \\
Abortion  & Women, Pro choice and Pro life\\
LGBTQ+ & 65 $\leq$ homosexual, bisexual, asexual, other (non heterosexual)\\
Healthcare & Have a chronic disease\\
Ageism & 65 $\leq$ Age $\leq$ 100\\
\bottomrule
\end{tabular}
\captionof{table}{Filters used in Prolific to allow participants in our studies. $*$ means we use only a subset of the basic set of residence countries (UK, US, Ireland, New Zealand, Australia), and $\dagger$ means we did not use English as first language. \label{tab:participants}}
\endgroup

\twocolumn

\clearpage

\section{Corpus Analysis}
\label{sec:appendix-corpus-analysis}

\begin{figure}[h]
\begin{tabular}{c}
\multicolumn{1}{c}{\textbf{Character X}}\\
\includegraphics[scale=.5]{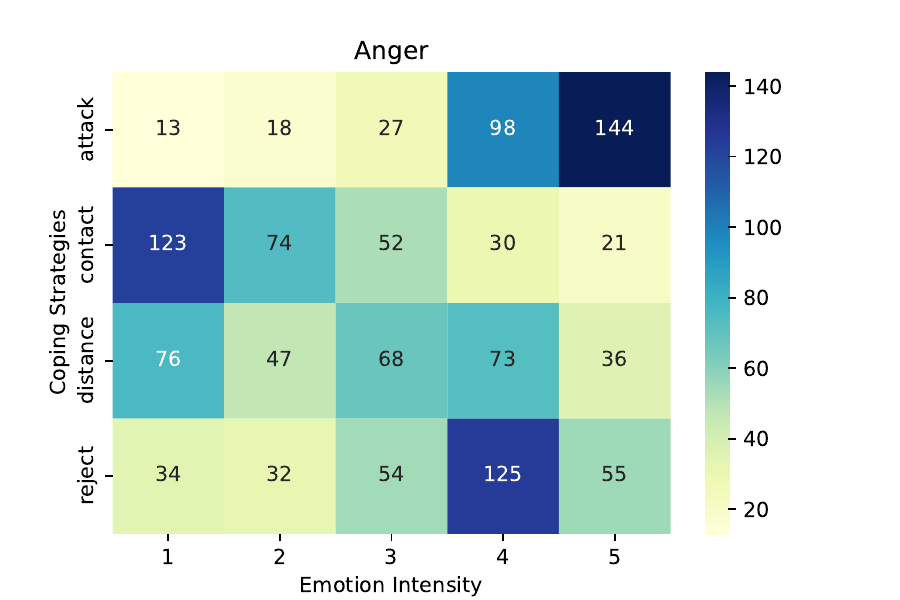}\\[5pt]

\multicolumn{1}{c}{\textbf{Respondent}}\\
\includegraphics[scale=.5]{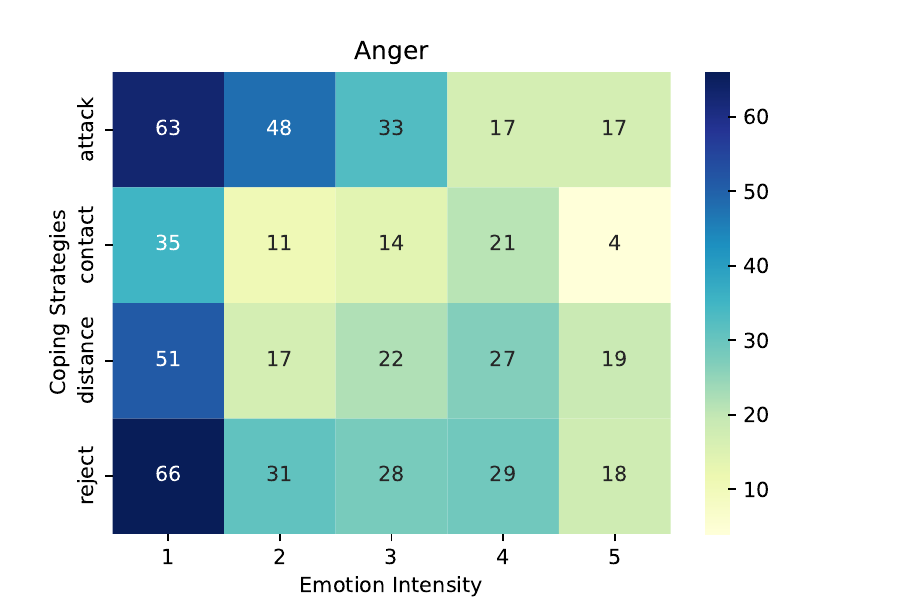}

\end{tabular}
\caption{Comparison of anger scores assigned to character $X$ and as self-assigned by participants who would not react as $X$. Numbers in each cell are the numbers of participants picking a given emotion score (intensity).}
\label{fig:emo-per-coping-generation} 
\end{figure}

Table~\ref{tab:most-com-words} reports the 10 words with the highest
tf-idf across coping strategies.
Figure~\ref{fig:emo-per-coping-generation} reports an example
comparison between the emotions assigned to character $X$ and the same
emotion for respondents who claimed their own reaction to $Y$ would be
different from $X$'s.

\begin{table}[h]
\centering\small
\begin{tabular}{p{1cm}p{5cm}}
\toprule
Attack & think, drugs, rights, people, don, women, right, abortion, disagree, say\\
Contact & people, think, drugs, believe, different, say, agree, way, good, don \\
Distance & think, people, drugs, don, say, want, disagree, agree, rights, life \\
Reject & don, think, people, drugs, rights, want, say, need, believe, unborn\\
\bottomrule
\end{tabular}
\caption{Most common words per coping strategy, using tf-idf (to be redone, this was not found on all data).}
\label{tab:most-com-words}
\end{table}

\vfill\eject 

\clearpage

\onecolumn
\section{Classification Experiments}
\label{sec:appendix-modeling}

We conduct zero-shot classification experiments on our local machines,
with the quantized models \textsc{phi-2.Q8\_0.gguf},
\textsc{mistral-7b-instruct-v0.2.Q8\_0.gguf},
\textsc{dolphin-2.0-mistral-7b.Q5\_K\_M.gguf}, and
\textsc{nous-hermes-2-solar-10.7b.Q5\_K\_M.gguf}, all downloaded from
\url{https://huggingface.co/TheBloke}.

Table~\ref{tab:zero-shot-llm} reports a structured difference between
the prompts (P1, P2, and P3) used in the zero-shot classification
experiments. Each prompt was used in three setups, i.e., by
concatenating the task instructions with $X$'s answer only, with the
description of $X$'s behavior only, or with a concatenation of $X$'s
answer and behavior.

\begin{table*}[h]
{\small
\begin{tabularx}{\textwidth}{p{.1cm}p{.4cm}p{4cm}p{4cm}p{5.5cm}}
\toprule
& & Prompt 1 & Prompt 2 & Prompt 3\\
\cmidrule(r){3-3}\cmidrule(r){4-4}\cmidrule(r){5-5}

\raisebox{-1.1cm}{\rotatebox{90}{Introduction}} && People respond to emotional situations with four different strategies. & People can oppose a view by adopting one of four emotional attitudes. & According to Roseman, there exist four emotion coping strategies, i.e., attack, contact, distance and reject.\\
\cmidrule(r){3-3}\cmidrule(r){4-4}\cmidrule(r){5-5}
\raisebox{-2cm}{\rotatebox{90}{Definitions}} &&Attack: the experiencer responds in an aggressive manner. Contact: the experiencer responds in an understanding manner. Distance: the experiencer seeks personal space away from the stimulus. Reject: the experiencer promptly refuses the stimulus. & Attack: to right the wrong in an aggressive manner. Contact: to right the wrong with an open mindset. Distance: to seek shelter from unpleasant situations in a quiet inner space. Reject: to approach life in a firm refusal of negativity. & Attack: coming across as aggressive and combative, facing conflicts fiercely, taking action to right the wrong. Contact: acknowledging the diversity of opinions, responding with a constructive attitude, trying to solve problems in a respectful manner. Distance: seeking emotional tranquillity and self-reflection, coming across as avoidant, disengaging from conflicts. Reject: judgmental attitude, resolute attempt to refuse unpleasant situations, expressing opinions with confidence, dismissing potential conflicts. \\
\cmidrule(r){3-3}\cmidrule(r){4-4}\cmidrule(r){5-5}
\multirow{3}{*}{\raisebox{-3.2cm}{\rotatebox{90}{Task}}} & & Please classify the following text with the one strategy being implemented in \ldots & Please classify the following text with the one attitude emerging from & Given these strategies, please recognize what strategy is being used by the \ldots \\\\
& \raisebox{-.6cm}{\rotatebox{90}{Answer}}& \ldots it (choose one and only one label). & \ldots it (choose one and only one label). & \ldots utterer of the following text (chose one and only one label). \ldots\\\\
& \raisebox{-1.1cm}{\rotatebox{90}{Behavior}} & \ldots the described behavior (choose one and only one label) -- note: the text is about how person X behaves in an unpleasant situation, using a coping strategy. & \ldots the described behavior (choose one and only one label) -- note: the text is about how person X behaves in an unpleasant situation, using a coping strategy. & \ldots person behaving as described in the following text (chose one and only one label) -- note: the text is about how person X behaves in an unpleasant situation, using a coping strategy.\\\\
& \raisebox{-1.6cm}{\rotatebox{90}{\makecell{Answer\\ and Behavior}}}& \ldots it (choose one and only one label) -- note: what follows includes a verbal reaction to an unpleasant situation and a description of how the utterer behaves in that context. & \ldots it (choose one and only one label) -- note: what follows includes a verbal reaction to an unpleasant situation and a description of how the utterer behaves in that context. & \ldots utterer of the following text (chose one and only one label) -- note: what follows includes a verbal reaction to an unpleasant situation and a description of how the utterer behaves in that context.\\

\bottomrule
\end{tabularx}
}
\caption{Zero-shot setup prompts, obtained by concatenating the Introduction, the coping strategies Definitions, and one labeling Task either focusing on $X$'s answer, $X$'s behavior or a combination of the two.}
\label{tab:zero-shot-llm}
\end{table*}

\clearpage 

Results obtained with all prompts are in Table~\ref{tab:llm-coping-classification}.
\F scores reported in Table~\ref{tab:Bert-and-others} correspond to P3.

\begin{table*}[h]
\centering\small
\setlength{\tabcolsep}{5pt}
\begin{tabular}{lccccccccc}
\toprule
&\multicolumn{3}{c}{\textbf{Answer}}&\multicolumn{3}{c}{\textbf{Behavior}}&\multicolumn{3}{c}{\textbf{Answ.+Behav.}}\\
\cmidrule(r){2-4}\cmidrule(r){5-7}\cmidrule(r){8-10}
& P1 & P2 & P3 & P1 & P2 & P3 & P1 & P2 & P3 \\
\cmidrule(r){2-2}\cmidrule(r){3-3}\cmidrule(r){4-4}\cmidrule(r){5-5}\cmidrule(r){6-6}\cmidrule(r){7-7}\cmidrule(r){8-8}\cmidrule(r){9-9}\cmidrule(r){10-10}
\textbf{phi-2} & .345 & .307 & .356& .338 & .350 &.391 & .397 & .321 & .322\\
\textbf{mistral-2} & .380 & .379 &.392 & .473 & .520 & .474 & .495 & .476 & .538\\
\textbf{dolphin-2} & .385 & .376 & .422 & .449 & .513 &{\bf .550} & .454 & .448 & .520\\
\textbf{nous-hermes} & .429 & .480 & .387 & .482 & .518 & .526 & 
.523 & .529 & .515\\
\bottomrule
\end{tabular}
\caption{\F with different prompts, with $X$'answer as input (Answer),
  with the description of $X$'s behavior (Behavior), or a
  concatenation of both (Answ.+Behav.). Each of these input types were
  separately concatenated with three prompts (P1, P2, and P3), fed to
  the four models (rows).}
\label{tab:llm-coping-classification}
\end{table*}

\end{document}